\documentclass{article} 
\usepackage[dvipsnames]{xcolor}
\usepackage{iclr2024_conference,times}

\usepackage[utf8]{inputenc} 
\usepackage[T1]{fontenc}    
\usepackage{hyperref}       
\usepackage{url}            
\usepackage{booktabs}       
\usepackage{amsfonts}       
\usepackage{nicefrac}       
\usepackage{microtype}      

\usepackage{amsmath}
\usepackage{amssymb}
\usepackage{mathtools, adjustbox}
\usepackage{amsthm}
\usepackage{dsfont}
\usepackage{float}
\usepackage{caption}
\usepackage{enumitem, wrapfig}

\usepackage{graphicx, placeins}
\usepackage{subcaption}
\usepackage{booktabs} 
\usepackage{multirow}
\usepackage{tabularx}

\usepackage{algorithmicx}
\usepackage[noend]{algpseudocode}
\usepackage{algorithm,algpseudocode}

\newcommand{\aug}[1]{\mathcal{A}(\cdot|#1)}
\newcommand{\augp}[2]{\mathcal{A}(#1|#2)}
\newcommand{\pndata}{\mathcal{P}_{\overline{\mathcal{X}}}}

\newcommand{\ndata}{\overline{\mathcal{X}}}
\newcommand{\adata}{\mathcal{X}}

\newcommand\R{\ensuremath{\mathbb{R}}}

\newcommand{\Exp}[1]{\mathrm{\mathbb{E}}_{#1}}
\newcommand{\wpair}[2]{w_{#1#2}}

\usepackage{pifont}

\newcommand{\con}{w}

\newcommand\ds[1]{\texttt{#1}}
\newcommand\red[1]{\textcolor{red}{#1}}
\newcommand\green[1]{\textcolor{ForestGreen}{#1}} 
\newcommand{\ours}{\textsc{LateTVG} } 
\newcommand{\al}{\alpha}
\newcommand{\be}{\beta}
\newcommand{\ga}{\gamma}
\newcommand{\ro}{\rho}

\theoremstyle{plain}
\newtheorem{theorem}{Theorem}[section]

\newtheorem{lemma}[theorem]{Lemma}
\newtheorem{corollary}[theorem]{Corollary}
\theoremstyle{definition}
\newtheorem{definition}[theorem]{Definition}
\newtheorem{assumption}[theorem]{Assumption}
\theoremstyle{remark}

\usepackage{titlesec}
\titlespacing*{\section}{0pt}{*1}{*1}

\title{Views Can Be Deceiving: Improved SSL Through Feature Space Augmentation}

\author{Kimia Hamidieh\textsuperscript{\rm 1} \quad Haoran Zhang\textsuperscript{\rm 1} \quad Swami Sankaranarayanan\textsuperscript{\rm 2} \quad Marzyeh Ghassemi\textsuperscript{\rm 1} \\
\textsuperscript{\rm 1}MIT,
\textsuperscript{\rm 2}Sony AI \\
\texttt{\{hamidieh,haoranz,swamiviv,mghassem\}@mit.edu}
}

\iclrfinalcopy 
\begin{document}

\maketitle
\begin{abstract}
Supervised learning methods have been found to exhibit inductive biases favoring simpler features. When such features are spuriously correlated with the label, this can result in suboptimal performance on minority subgroups.
Despite the growing popularity of methods which learn from \textit{unlabeled} data, the extent to which these representations encode spurious features is unclear. In this work, we explore the impact of spurious features on Self-Supervised Learning (SSL) for visual representation learning. We first empirically show that commonly used augmentations in SSL can cause undesired invariances in the image space, and illustrate this with a simple example. We further show that classical approaches in combating spurious correlations, such as dataset re-sampling during SSL, do not consistently lead to invariant representations. 
Motivated by these findings, we propose \ours to remove spurious information from these representations during pretraining, by regularizing later layers of the encoder via pruning. 
We find that our method produces representations which outperform the baselines on several benchmarks, without the need for group or label information during SSL.
\end{abstract}

\section{Introduction}
Standard supervised machine learning models exhibit high overall performance but often perform poorly on minority subgroups~\citep{shah2020pitfalls,mccoy2019right,gururangan2018annotation}. One potential cause is the presence of spurious correlations, \mbox{which are features that are only correlated with the label for specific subsets of data.} For instance, a machine learning model tasked with predicting bird species from images across different habitats may use the background the bird commonly appears in as a ``shortcut'', instead of core features specific to the bird such as the shape of their beak or plumage. This results in poor performance on bird groups that appear in unexpected environments~\citep{sagawa2020groupdro}. 
Identifying spurious correlations in the supervised learning setting has been well studied, where empirical risk minimization has been shown to exploit spurious correlations and result in poor performance for minority subgroups~\citep{hashimoto2018fairness}. As downstream tasks are explicitly defined, the label can be used to distinguish between core and spurious features~\citep{liu2021just, zhang2022correct}. Recent work has proposed various methods to identify and mitigate the effects of spurious features, such as learning multiple prediction heads~\citep{lee2022diversify}, causal inference~\citep{creager2021environment}, data augmentation~\citep{gao2023out} and targeted strategies such as importance weighting~\citep{lahoti2020fairness}, re-sampling~\cite{idrissi2021simple, tu2020empirical}, or approaches based on group distributionally robust optimization~\citep{sagawa2020groupdro,duchi2019distributionally}.

More recently, self-supervised learning (SSL) has emerged as a common form of pre-training for task-agnostic learning with large, unlabeled datasets~\citep{chen2020simple,moco,byol,simsiam,swav,zbontar2021barlow, chen2020improved}. SSL methods learn representations from unlabeled datasets by solving an auxiliary pretext task~\citep{doersch2015unsupervised}, such as inducing invariance between the representations
of two augmented views of the same image~\citep{moco, chen2020simple}. 
These methods have shown impressive results for a wide range of downstream tasks and datasets~\citep{liu2021self, jaiswal2020survey, tamkin2021dabs}.

Capturing \emph{core} features -- rather than spurious features -- is essential for learning effective representations that can be used in downstream tasks, but is particularly difficult in the case of SSL due to the absence of labeled data during the pre-training process.
Given only unlabeled data, we define spurious features as those that strongly correlate with core features for most examples in the training set, but are not useful for downstream tasks. For example, when training an SSL model on multi-object images, larger objects may interfere with the learning of smaller objects~\citep{chen2021intriguing}. If the downstream task involves only the prediction of smaller objects, the larger (spurious) object may suppress the smaller (core) object from being learned.
Large-scale unlabeled datasets that are commonly used in machine learning are inevitably imbalanced~\citep{van2021benchmarking}, have been found to be biased towards spuriously correlated sensitive attributes~\citep{calude2017deluge} such as gender or race~\citep{agarwal2021evaluating}, and can also include 
label-irrelevant features~\citep{torralba2011unbiased, fan2014challenges}. 

In this paper, we investigate the impact of spurious correlations on SSL pre-training. 
We first show theoretically that image augmentations used in SSL pre-training can lead to spurious connectivity when learning representations, causing the model to fail to predict the label using core features in downstream tasks. We empirically evaluate spurious connectivity, and then show that existing methods for utilizing group information in ERM based approaches do not provide an analogous improvement in SSL pre-training.
We then propose \emph{\textbf{Late}-layer \textbf{T}ransformation-based \textbf{V}iew \textbf{G}eneration} or \ours -- a method that induces invariance to spurious features in the representation space by regularizing final layers of the featurizer via pruning. 
Importantly, since our approach addresses SSL pre-training, we do not assume that model developers know \textit{apriori} the identity or values of the spurious features that exist in the data. We first evaluate \ours on several popular benchmarks for spurious feature learning, and then connect our method to the theoretical analysis by showing that \ours models empirically exhibit lower spurious connectivity. Our method demonstrates improved discriminative ability, especially over minority subgroups, for downstream predictive tasks, without access to group or label information. 
We make the following contributions:

\vspace{-0.2cm}
\begin{itemize}
    \item We provide theoretical arguments (Sec~\ref{sec:analysis}) that illustrate how common augmentations used in SSL pre-training affect the model's ability to rely on spurious features, for downstream linear classifiers.
    \vspace{-0.1cm}
    \item We explore the extent of spurious learning in self-supervised representations through the lens of downstream worst-group performance. We empirically show that known techniques for avoiding spurious correlations, such as re-sampling of the training set given group information, do not consistently improve core feature representations (Sec~\ref{sec:resample}). 
    \vspace{-0.1cm}
    \item 
    We propose \ours -- an approach that corrects for the biases caused by augmentations, by modifying views of samples in the representation space (Sec~\ref{sec:method}). 
    We find that \ours effectively improves worst-group performance in downstream tasks on four datasets by enforcing core feature learning (Sec~\ref{sec:results}).
\end{itemize}

\section{Related Work} 
\paragraph{Spurious Correlations.} Spurious correlations arise in supervised learning models~\cite{koh2021wilds, joshi2023towards, singla2021salient} in a variety of domains, from medical imaging \citep{zech2018variable, degrave2021ai} to natural language processing \citep{tu2020empirical, wang2020identifying}. A variety of approaches have been proposed to learn classifiers which do not make use of spurious information. Methods like GroupDRO~\citep{sagawa2020groupdro} and DFR \citep{kirichenko2022last} require group information during training, while methods like JTT~\citep{liu2021just}, LfF \citep{nam2020learning}, CVaR DRO \citep{duchi2019distributionally}, and CnC \citep{zhang2022correct} do not. However, all methods require group information for model selection. 
\vspace{-0.1in}
\paragraph{Self-supervised Representation Learning.} 
Self-supervised learning methods learn representations from large-scale unlabeled datasets where annotations are scarce.
In vision applications, the pretext task is typically to maximize similarity between
two augmented views of the same image \citep{jing2020self}. This can be done in a contrastive fashion using the InfoNCE loss \citep{oord2018representation}, such as in \citet{chen2020simple} and \citet{chen2020improved}, or without the need for negative samples at all, as in ~\citet{byol, swav, simsiam, dino,oquab2023dinov2, zbontar2021barlow}. Prior work has shown that SSL models may learn to spuriously associate certain foreground items with certain backgrounds \citep{meehan2023ssl}, In this work, we explore one potential mechanism for this phenomenon, both theoretically and empirically.

\vspace{-0.1in}
\paragraph{Representation Learning under Dataset Imbalance and Shortcuts.}
Self-supervised models have demonstrated increased robustness to dataset imbalance \citep{liu2021self, jiang2021self,jiang2021improving}, and the dominance of easier or larger features suppressing the learning of other features \citep{chen2021intriguing}. Some prior work has addressed shortcut learning in contrastive learning through adversarial feature modification without group labels \citep{robinson2021can}. However, other approaches to group robustness or fairness in self-supervised learning require group information or labels \citep{tsai2020demystifying, song2019learning,wang2021self, bordes2023guillotine, scalbert2023improving}. This paper focuses on learning representations from an unlabeled dataset with spurious correlations, encompassing both dataset imbalance and features of varying difficulty.
\vspace{-0.1in}
\paragraph{Regularization in Self-supervised Learning.}
The concept of regularizing a specific subset of the network is relatively unexplored in self-supervised learning but finds motivation in recent findings from supervised settings, such as addressing minority examples \citep{hooker2019compressed}, out-of-distribution generalization \citep{zhang2021can}, late-layer regularizations through head weight-decay \citep{abnar2021exploring}, and initialization \citep{zhou2022fortuitous}. Additionally, \citet{lee2022surgical} propose surgically fine-tuning specific layers of the network to handle distribution shifts in particular categories. These studies provide support for the approach of targeting a specific component of the network in self-supervised learning.

\section{Spurious Connectivity Induces Downstream Failures}
\label{sec:spurious_downstream_failures}

In this section, we introduce a toy setting to demonstrate that common augmentations used in SSL pre-training affect a model's ability to rely on spurious features for downstream linear classifiers. We consider a binary classification problem with a binary spurious attribute, with an equal number of samples per group~(Section \ref{sec:toy}). 
We show that augmentations applied during SSL pre-training can introduce undesired invariances in the representation space learned by a contrastive objective, making the downstream linear classifier trained on representations more reliant on the spurious feature~(Section \ref{sec:analysis}).


\subsection{Background and Setup}
\begin{figure}[tbp]
    \centering
    \vspace{-0.5in}
    \begin{subfigure}{0.35\textwidth}
        \includegraphics[width=\linewidth]{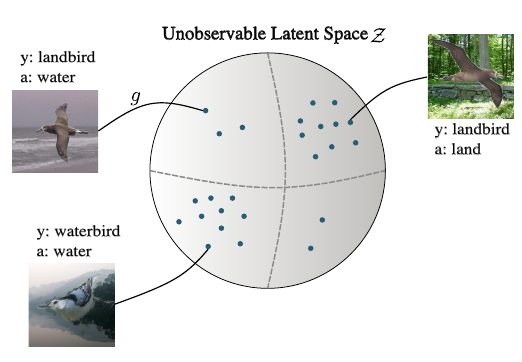}
        \caption{}
        \label{fig:intro}
    \end{subfigure}
    \hfill
    \begin{subfigure}{0.60\textwidth}
        \includegraphics[width=\linewidth]{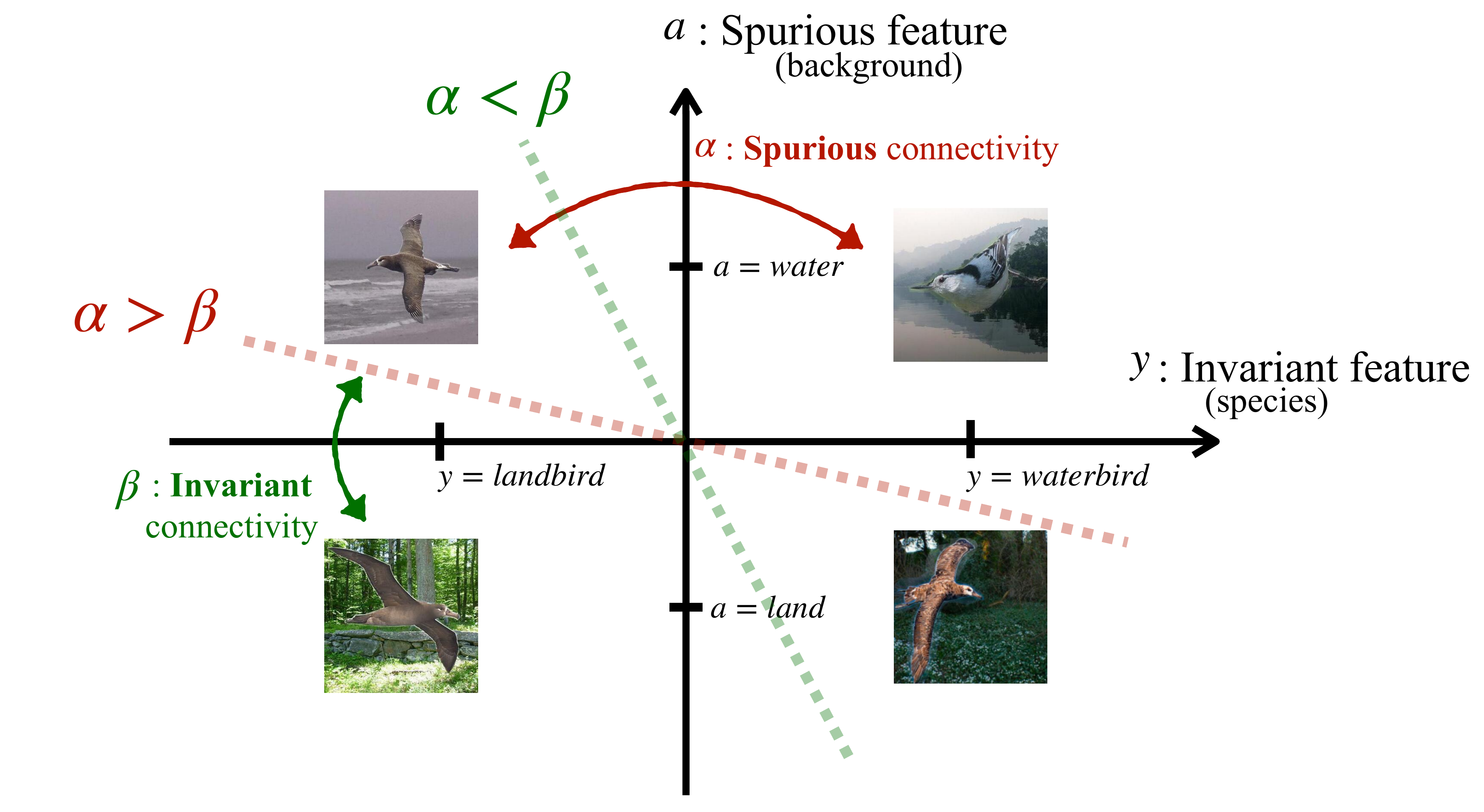}
        \caption{}
         \label{fig:analysisb}
    \end{subfigure}
    \caption{\textbf{Analysing SSL augmentations}. (a) Images generated from a latent space with correlating features. (b) If the connectivity induced by SSL augmentations between subgroups with the same spurious features is higher than the ones with the same invariant features, learned representations lead a downstream linear model to separate the data based on the spurious feature (red dashed line) instead of the invariant feature (green dashed line). Our empirical evaluation in Table~\ref{tab:connectivity} shows that this is indeed the case across different datasets considered in this work.
    \vspace{-1em}
    }
    \label{fig:analysis}
\end{figure}

\paragraph{Setup. } 
We consider learning representations from an unlabeled data space $\adata$ generated from an underlying latent feature space 
$\mathcal{Z} \in \mathbb{R}^m := \{z_\text{core}, z_\text{spur}, \dots, z_m\}$, 
where $z_\text{core}$ and $z_\text{spur}$ are correlated features.
For a given downstream task with labeled samples, we assume that each $x\in \adata$ belongs to a class given by the ground-truth labeling function $y: \adata\rightarrow \mathcal{Y}$ where $z_\text{core}$ determines the labels for our downstream task of interest, while $z_\text{spur}$ determines the spurious attribute, which is easier to learn, and is not of interest for downstream tasks. We can define a deterministic attribute function $a: \adata\rightarrow \mathcal{S}$ where each $x \in \adata$ takes a value in $\mathcal{S}$. 
Let $g = (y(x), a(x))$ denote the subgroup of a given sample $x$, where $\mathcal{G} = \mathcal{Y} \times \mathcal{S}$ is the set of all possible subgroups. Figure~\ref{fig:analysis} illustrates the subgroups on the Waterbirds dataset, where the background is a spurious feature that correlates with the bird species.


\paragraph{Contrastive learning. } We aim to learn representations by bringing together data-augmented views of the same input, which we refer to as positive pairs, using a contrastive objective. Let $P_+$ be the distribution of positive pairs, which can be defined as the marginal probability of generating the augmented pair $x$ and $x'$ from the same image in the (natural) population data. Thus the distribution $P_+$ relies both on original data distribution and the choice of SSL augmentations. 
To analyze the representation space learned in contrastive learning and core feature predictivity of the representations, consider a weighted graph with vertex set $\adata$ where the undirected edge $(x, x')$ has weight $w_{xx'} = P_+ (x, x')$ similar to augmentation graph in~\citet{haochen2021provable}.

Although the augmentation graph learns semantically similar structures that enables generalization to new domains~\citep{shen2022connect}, the inductive biases set by these augmentations is not well studied. 
In this work, we draw attention to cases where augmentations can create \emph{spurious connectivities} within subgroups of the data, and when and why these connectivities can cause the downstream linear model to rely on the spurious feature. 

\subsection{Spurious Connectivity in a Toy Setup}
\label{sec:toy}
In this section, we introduce a setting in which contrastive objectives can learn representations that cause linear downstream models fail on downstream tasks. 
To start, we investigate how augmentations can transform the samples such that the subgroup assignment changes. 

\begin{definition} \emph{Subgroup connectivity. }
\label{def:con}
Define the average subgroup connectivity given two disjoint subsets $G_1, G_2 \subseteq \adata$ as $    \con(G_1, G_2) = \frac{1}{|G_1|.|G_2|} \sum_{x\in G_1, x' \in G_2} w_{xx'}$.  
where $w_{xx'}$ is the probability of generating the augmented pair $x$ and $x'$ from the same image in the natural population data. 
\end{definition}

Intuitively, this subgroup connectivity is the average weight of edges connecting $G_1$ to $G_2$, and is proportional to the probability of a sample $x\in G_1$ being transformed to a sample $x'\in G_2$ via augmentations. See Appendix \ref*{app:theory} for further details. 

We specifically define the following terms to be the expected value of $w(G_1, G_2)$ from Definition~\ref{def:con}, when subgroups $G_1$ and $G_2$ have the following properties: 
\begin{itemize}[itemsep=0mm]
    \item \textbf{Spurious connectivity ($\al$)}: $G_1$ and $G_2$ share the same spurious attribute but differ in class 
    \item \textbf{Invariant connectivity ($\be$)}:  $G_1$ and $G_2$  share the same class but differ in spurious attribute %
    \item \textbf{Opposite connectivity ($\ga$)}:  $G_1$ and $G_2$  differ both in the spurious attribute and the label 
\end{itemize}

Where $\al$, $\be$, $\ga$ are average values estimated across a dataset consisting of subgroups. 

\paragraph{Toy Setup. } 
\label{def:toysetup} 
\newcommand{\toy}{\text{toy}}

We consider a downstream classification problem where a spurious attribute is present, and both the input and the spurious attribute take binary values. We define the probability of sampling a positive pair $(x,x')$ based on the expected connectivity terms 
$\alpha_{\toy}$, $\beta_{\toy}$, $\gamma_{\toy}$, and $\rho_{\toy}$ as follows:

\vspace{-1em}
\begin{equation*}
P_+(x, x') = \begin{cases}
\alpha_\toy, &\text{if } a(x) \neq a(x') \text{ and } y(x) = y(x') \\
\beta_\toy, &\text{if } a(x) = a(x') \text{ and } y(x) \neq y(x') \\
\gamma_\toy, &\text{if } a(x) \neq a(x') \text{ and } y(x) \neq y(x') \\
\rho_\toy, &\text{if } a(x) = a(x') \text{ and } y(x) = y(x')
\end{cases}
\label{eq:positive-pair-prob}
\end{equation*}

Note that the average subgroup connectivity for this setup, would be exactly the same as the corresponding connectivity variable. Thus in our running example we have $\al = \al_{\toy}, \be = \be_{\toy}, \ga = \ga_{\toy}$, and we can use them interchangeably. 
For this simplified augmentation graph, the expected connectivity terms between groups are a property of the graph, and independent of the model or architecture we use for learning representations. Combined with a contrastive objective, the expected connectivity can be a proxy for how close different subgroups are going to be in the representation space.

\subsection{Analysis of the Toy Setting}
\label{sec:analysis}

In Section~\ref{sec:empasmpt}, we empirically show that common augmentations used in contrastive learning can be detrimental to learning invariant representations, as they implicitly encourage samples to cluster primarily based on the spurious feature. 
Based on this observation, we make the following assumption. 

\begin{assumption}\label{assumption:distortion} 
	Given a spurious attribute function $a: \adata\rightarrow |G| $ which is defined for all $x \in \adata$, we assume that for a data point $x \in \adata$, the probability of distorting the labeling of the augmented images sampled from the augmentation distribution $\aug{\bar{x}}$, is greater than the probability of distorting the attribute. More formally,
 \begin{align*}
     \Pr_{\tilde{x}\sim \aug{x}} &\left({y}(\tilde{x}) \ne y(x), {a}(\tilde{x})= a(x) \right ) \geq  \Pr_{\tilde{x}\sim \aug{x}} \left({y}(\tilde{x})= y(x), {a}(\tilde{x})\ne a(x)\right)
 \end{align*}
\end{assumption}

\begin{lemma}\label{lemma:main}
Consider the set of (unlabeled) population data $\adata$ in a binary-class setting where the spurious attribute takes binary values, consisting of $|\mathcal{G}| = 4$ groups, with the same number of examples per group. Consider a simplified augmentation graph with parameters $\al$, $\be$, $\ro$, $\ga$ defined as in ~\ref{def:toysetup}, and assume that augmentations are more likely to change either class or attribute, than to change neither of the two ($\al > \ga, \be > \ga$), and that augmentations are less likely to change both at the same time ($\ro > \al, \ro > \be$).  

Under these conditions, the spectral contrastive loss recovers both invariant and spurious features, and for each sample in the population data, the spurious feature is bounded by constant $B_{sp} = \sqrt{\be - \al - \ga + \ro}$, while the invariant feature is bounded by $B_{inv} = \sqrt{\al - \be - \ga + \ro}$, in the representation space. \text{Proof in Appendix~\ref*{appx:proof}.}
\end{lemma}

\begin{corollary}
Given Assumption~\ref{assumption:distortion}, where $\al > \be$ in the simplified augmentation graph, the margin of the spurious classifier is $B_{sp}$, and is less than the margin of the invariant classifier $B_{inv} $, and the max-margin classifier trained on representations given by spectral clustering converges to the spurious classifier. 
\end{corollary}

This suggests that even with the same number of samples across different groups during pre-training, downstream linear classifiers can rely on the spurious feature to make predictions, where the representations are determined by the simplified augmentation graph and the spectral contrastive loss.

\section{Exploring Spurious Learning in Representations}\label{sec:explore_spur}

In this section, we investigate the performance of downstream linear models trained on self-supervised representations, empirically verify our assumption regarding spurious and invariant connectivity, and show that in practice -- similar to our toy analysis -- having the same number of examples across groups in the presence of spurious connectivity does not lead to performance gains. 

\vspace{-0.1in}
\subsection{Experimental Setup}\label{subsec:exp-setup}
\vspace{-0.1in}
\paragraph{Datasets} We evaluate methods on five commonly used benchmarks in spurious correlations -- CelebA~\citep{liu2015deep}, CMNIST~\citep{arjovsky2019invariant}, MetaShift~\citep{liang2022metashift}, Spurious CIFAR-10~\citep{nagarajan2020understanding}, and Waterbirds~\citep{wah2011caltech} (See Appendix \ref*{app:datasets} for dataset descriptions). For each dataset, we train an encoder with an SSL-based pre-training step followed by a supervised training of a linear model that probes the representations learned using SSL for the downstream task. 

\vspace{-0.1in}
\paragraph{SSL Pre-training} For the SSL pre-training, we train SimSiam~\citep{simsiam} models with a ResNet backbone throughout the paper. The training split used during the pre-training stage are unbalanced and contain spuriously correlated data. The group/label counts for each dataset and split is shown in Appendix~\ref*{app:datasets}. The backbone network used for most of our experiments are initialized with random weights, unless specified otherwise. We additionally report results for SimCLR~\citep{chen2020simple} models in Section~\ref{sec:ours-res}, 

\vspace{-0.1in}
\paragraph{Downstream Task} For downstream task prediction, we train a linear layer using logistic regression on top of the pretrained embeddings. Note that the backbone is frozen during this finetuning phase and only the linear layer is updated. We use a balanced dataset for training where the spurious correlation does not hold. To create this downstream training dataset, we subsample majority groups~\citep{sagawa2020investigation,idrissi2021simple}, to avoid the geometrical skews~\citep{nagarajan2020understanding} of the linear classifier on representations. Then, we evaluate the learned representations on the standard test split of each dataset, where group information is given. For each run, we report the average and worst-group accuracy. 

\vspace{-0.1in}
\paragraph {Empirical Evaluation of Spurious Connectivity}
\label{sec:assumpt}

To evaluate the connectivity term for each pair of subgroups in datasets exhibiting spurious correlations, we conduct an empirical analysis similar to~\citet{shen2022connect}. Specifically, we train a classifier to distinguish between each pair of subgroups and evaluate its performance on a subset of the data that has been augmented with SSL augmentations. The error of the classifier represents the probability that the augmentation module alters the subgroup assignment for each example between the two subgroups, making them indistinguishable. Figure~\ref{fig:analysis} illustrates this procedure.

\vspace{-0.1in}
\paragraph{The Role of Initialization} In representation learning, encoders are not typically trained from scratch but initialized from a model pretrained on larger datasets, such as ImageNet \citep{deng2009imagenet}. Recent work in transfer learning~\citep{geirhos2018imagenet, salman2022does} has questioned this assumption and pointed out that biases in pretrained models linger even after finetuning on downstream target tasks. In this section and more broadly in our work, we focus on performing SSL pre-training from randomly initialized weights. In addition, since the datasets considered in this work are similar to ImageNet, the performance of off-the-shelf ImageNet pretrained models is expected to be higher. For completeness, we have added these results to Appendix~\ref*{app:imagenet}.
\vspace{-0.1in}
\subsection{High Levels of Spurious Connectivity in Practice}
\label{sec:empasmpt}
We measure connectivity across four datasets in Table~\ref{tab:connectivity}, and on all of them, we find that the average spurious connectivity is higher than invariant connectivity. We also confirm that both these values are higher than the probability of simultaneously changing both spurious attributes and invariant attributes. 
This means that the samples within the training set are more likely to be connected to each other \emph{through} the spurious attribute, rather than the core feature. This suggests that the contrastive loss prefers alignment based on the spurious attribute instead of the class. 

\begin{table}[htbp]
\caption{We report the error of classifiers trained to distinguish between two subgroups as a proxy for the probability of augmentations flipping group assignments between each two groups in the dataset, or the connectivity of two subgroups in the image space.}
\label{tab:connectivity}
\resizebox{0.6\linewidth}{!}{
\begin{tabular}{@{}cccc@{}}
\toprule
\textbf{Dataset}    & \multicolumn{1}{c}{\textbf{\begin{tabular}[c]{@{}c@{}}Spurious \\ Connectivity\end{tabular}}} & \multicolumn{1}{c}{\textbf{\begin{tabular}[c]{@{}c@{}}Invariant \\ Connectivity\end{tabular}}} & \multicolumn{1}{c}{\textbf{\begin{tabular}[c]{@{}c@{}}Opposite \\ Connectivity\end{tabular}}} \\ \midrule
\ds{celebA}     & 10.4                  & 3.7                   & 2.8 \\
\ds{cmnist}     & 31.6                  & 8.3                   & 6.8 \\
\ds{metashift}  & 16.3                  & 13.6                  & 5.0                            \\
\ds{waterbirds} & 25.3                  & 11.2                   & 7.8                            \\ \bottomrule

\end{tabular}
}
\centering
\end{table}

We compute the connectivity terms by training classifiers to distinguish augmented data from each combination of the two groups in the dataset and reporting their error rates. 

The details of the choice of augmentations and training for this step can be found in Appendix~\ref*{app:spconexpt}.

\vspace{-0.1in}
\subsection{SSL Models Learn Spurious Features}

To measure the reliance of downstream models to spurious correlations, we measure the accuracy of the downstream model on each group in the test set, and use the worst-performing group accuracy as a lens to reason about spurious correlations. 
We find across all datasets, SSL models exhibit gaps between worst-group and average accuracy when predicting the core feature (Table \ref{table:regularization} in Appendix~\ref*{app:reg}).  

These results indicate, that unlike supervised learning~\citep{menon2021overparameterisation,kirichenko2022last,rosenfeld2022domain}, training of the final layer on a balanced set where the spurious correlation does not hold is not sufficient for improving worst-group accuracy when predicting the core attribute.

\vspace{-0.1in}
\subsection{Resampling During SSL does not Improve Downstream Performance}
\label{sec:resample}
To probe the effect of availability of such group information during the SSL pre-training stage, we examine whether classical approaches for combating spurious correlations, such as re-sampling training examples~\citep{idrissi2021simple}, are effective in removing spurious information during SSL pre-training. 

Assuming that group information is available, we train SimSiam on datasets re-sampled using the following strategies: (i) \textit{Balancing} groups by resampling training examples to match the downstream validation distribution. (ii) \textit{Downsampling} examples in majority groups to have the same number of examples in all groups. (iii) \textit{Upsampling} minority examples to have the same number of examples in all groups.

\begin{table*}[!h]
\centering
\caption{\textbf{Worst-group accuracy difference} (\%) between each balancing strategy and the original training set. Original training performance are shown in parentheses below each dataset. Full results can be found in Appendix Table \ref{tab:sampling_strat_full}.}
\label{tab:resampling}
\resizebox{0.8\linewidth}{!}{
\begin{tabular}{@{}lrrrrr@{}}
\toprule
\textbf{Sampling Strategy} & \begin{tabular}[c]{@{}r@{}}\ds{celebA} \\ (77.5)\end{tabular} & \begin{tabular}[c]{@{}r@{}}\ds{cmnist} \\ (75.4)\end{tabular} & \begin{tabular}[c]{@{}r@{}}\ds{metashift}\\ (42.3)\end{tabular} & \begin{tabular}[c]{@{}r@{}}\ds{spurcifar10}\\ (43.4)\end{tabular} & \begin{tabular}[c]{@{}r@{}}\ds{waterbirds}\\ (48.3)\end{tabular} \\ \midrule

\textbf{Balancing}         & \red{-1.7}                                                                         & \red{-8.7}                                                                         & \red{-3.8}                                                                           & \red{-8.3}                                                                             & \green{+3.0}                                                                            \\
\textbf{Downsampling}      & \green{+0.3}                                                                         & \red{-10.6}                                                                        & \green{+3.9}                                                                           & \red{-14.4}                                                                            & \green{+0.5}                                                                            \\
\textbf{Upsampling}        &                             \green{+4.1}                                               & \red{-5.3}                                                                         & \green{+2.7}                                                                           & \red{-19.4}                                                                            & \red{-0.3}                                                                            \\ \bottomrule
\end{tabular}
}
\end{table*}

We find that re-sampling during self-supervised pre-training does not improve downstream worst-group accuracy in a consistent manner as in Table~\ref{tab:resampling}. We do see minor improvements for \ds{metashift} and \ds{celebA}, but contrast this with large drops for \ds{spurcifar10} and \ds{cmnist}. Given that the downstream linear model is trained on a downsampled dataset where such correlations do not exist, this means that re-sampling during self-supervised training does not necessarily improve the linear separability of representations with respect to the core feature, even given a balanced finetuning dataset. This is analogous to our findings in the toy setting in Section~\ref{sec:analysis}.

\section{Creating Robust Representations via Feature Space Augmentations}

In the previous sections, we showed that augmentation mechanisms used in SSL result in poor performance under spuriously correlated features in the training set. Instead of curating specific image augmentations that correct for these biases in the image space, we propose an approach to target spurious connectivity in the \textit{representation} space by modifying positive pairs. In this section, we describe our approach, \ours that improves the performance of SSL models by introducing pruning based regularization to the later layers of the encoder.

\vspace{-0.1in}

\subsection{\textbf{Late}-layer \textbf{T}ransformation-based \textbf{V}iew \textbf{G}eneration }
\label{sec:late_tvg}

Motivated by improved SSL model invariance when trained with augmentations in \emph{image} space~\citep{chen2020simple}, we propose a model transformation module that specifically targets augmentations that modify the spurious feature in \emph{representation} space. We propose \emph{Late-layer \textbf{T}ransformation-based \textbf{V}iew \textbf{G}eneration} - \ours, which uses feature space transformations to mitigate spurious learning in SSL models and improve learning of the core feature.

\label{sec:method}

Formally, we propose using a model transformation module $\mathcal{U}$, that transforms any given model $f_\theta$
parameterized by $\theta = \{W_1, \dots, W_n\}$ 
to 
$f_{\tilde{\theta}}$. At each step, we draw a transformation $\phi_{M, \theta} \sim \mathcal{U}$ 
to obtain the transformed encoder. 
 Each model transformation can be defined with a mask $M \in \{0, 1\}^{|\theta|} $, where we transform the unmasked weights $(1-M) \odot \theta$ by $\phi$, and keep the rest of the weights  $M \odot \theta$  the same to obtain $\tilde{\theta}$. Here, we propose a specific transformation module $\mathcal{U}$. 
\vspace{-0.1in}
\paragraph{Transformations. } For mitigating spurious connectivity, we choose a simple transformation targeted towards regularizing the final layers of the encoder. 
In our experiments, we consider a threshold pruning transformation module, which uses magnitude pruning on $a\%$ of the weights in all layers deeper than $L$. 
More specifically, we propose a model transformation module $\mathcal{U}_\text{\;Prune, L, a}$, with $\phi(\theta) = 0$,  $M := M_{L, a} = \{M^l_L \odot \text{Top}_{a}(W_l) \mid  l \in [n]\}$ and  
    $\text{Top}_{a}(W_l)_{i, j} = \mathbb{I}(\lvert {W_{l_{(i, j)}} \lvert } \text{ in top } a\% \text{ of } \theta) $. 
Note that in this specific setting, the module transformation is deterministic (i.e. $|\mathcal{U}| = 1$), though our formalization also allows for random transformations such as randomized pruning or re-initialization. 

\begin{figure*}[t]
\centering
\vspace{-1.5em}
\includegraphics[width=0.8\textwidth]{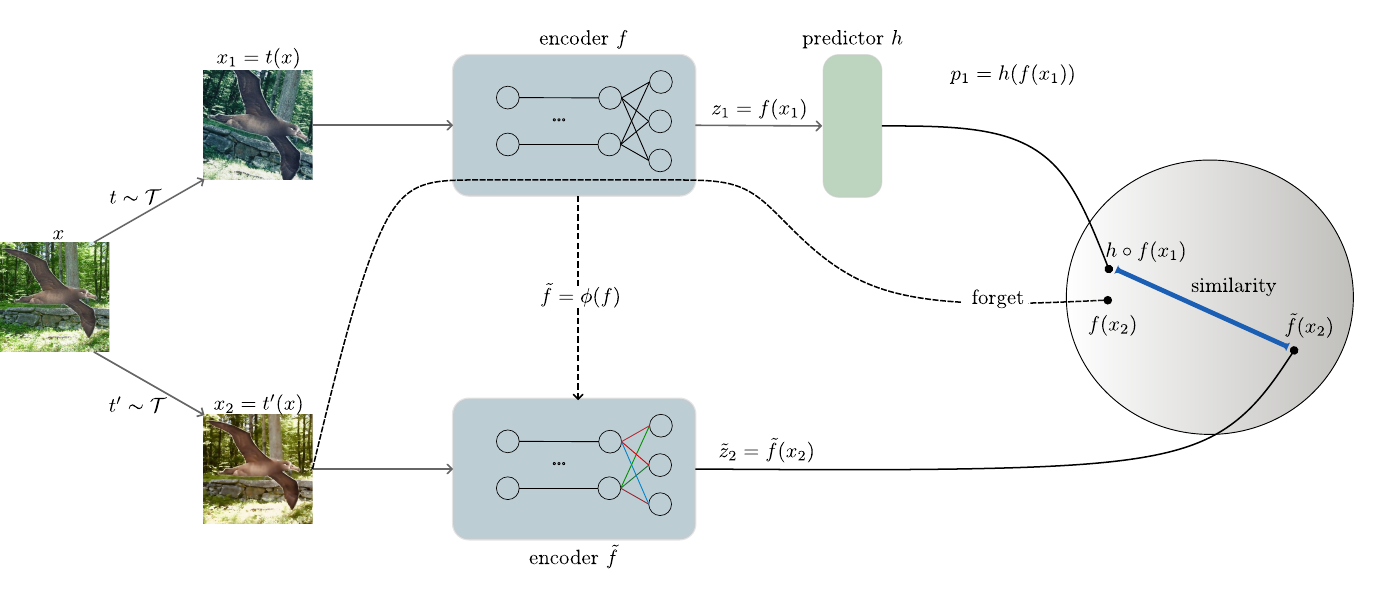}
\vspace{-.5em}
\caption{
We use model transformation modules to create new views of training examples in the representation space. The introduced set of transformations removes the features learned in the final few layers, and provides final representations invariant to such transformations. 
\label{fig:main}
\vspace{-.8em}
}
\end{figure*}

To learn these representations, given two random augmentations $t, t' \sim \mathcal{T}$ from the augmentation module $\mathcal{T}$, two views $x_1 = t(x)$ and $x_2 = t'(x)$ are generated from an input image $x$. At each step, given a feature encoder $f$, and an augmentation module $\mathcal{U}$, we obtain a transformed model $\tilde{f} = \phi(f)$ with $ \phi \sim \mathcal{U}$. 
During training, examples $x_1$ and $x_2$ are respectively passed through the normal encoder $v_1 = f(x_1)$, and the transformed encoder $\tilde{v}_2 = \tilde{f}(x_2)$. Encoded feature $\tilde{v}_2$ is now a positive example that should be close to $v_1$ in the representation space. An algorithmic representation of the method can be found in Appendix~\ref{app:alg}. 

\vspace{-0.1in}
\paragraph{Intuition for LateTVG. } When learning a discriminative process that maps data to a separable space, the variance among different subpopulations is stored in distinct regions of the network~\citep{lee2022surgical}. As a result, both spurious and core features, which describe the high-level data distribution, tend to reside at the end of a neural network. Thus, in \ours, we aim to encourage the final layers to learn more difficult features, by applying a model transformation that targets these layers, and causing the model to be invariant to final layer transformations. As pruning in supervised models have been shown to affect minority examples more than majority ones \citep{hooker2019compressed}, we hypothesize that our transformation can be considered as a curated view generating operation for the minority groups. In particular, pruning would contribute to ``forgetting'' the minority examples from the network, resulting in upweighting the loss for these examples. 

\vspace{-0.1in}

\subsection{Experiments}
\label{sec:results}

In this section, we demonstrate the efficacy of \ours in mitigating the dependence on spurious correlations. We use the same experimental setup as described in Section~\ref{subsec:exp-setup}. 
For evaluation of \ours, we use our SSL-\ours approach during the pre-training stage. We compare this performance to SSL models pre-trained with the standard SSL-base trained with either SimSiam or SimCLR.

\vspace{-0.1in}
\subsubsection{\ours Improves SSL Worst-Group Performance}
\label{sec:ours-res}

\begin{wrapfigure}{r}{0.45\textwidth}
    \centering    
    \includegraphics[width=0.5\textwidth]{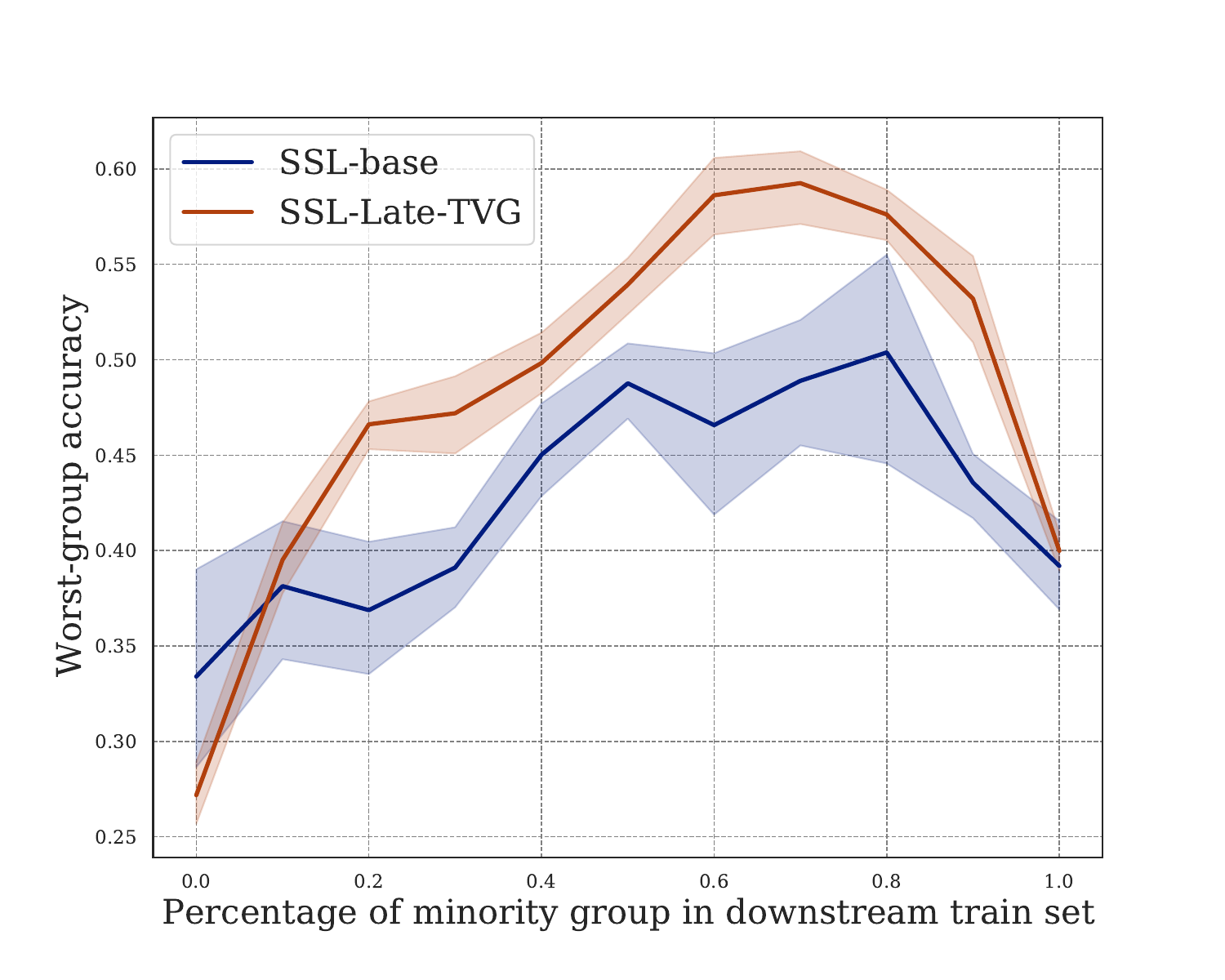}
    \caption{Downstream worst-group accuracy of SSL-Late-TVG on the \ds{metashift} dataset as we vary the percentage of minority group in the downstream training set. For all cases except for extreme minority decrement, SSL-Late-TVG outperforms the baseline. }
    \label{fig:minorchange}
\end{wrapfigure}

The goal of this experiment is to understand how \ours affects worst-group performance in downstream tasks that use SSL representations. We compare the worst group accuracy of two approaches, SSL-Base and SSL-\ours on 5 different datasets. Both models used similar hyper-parameter grids and model selection criteria as noted previously. The results are presented in Table \ref{tab:prune-all}. We show the performance of the best hyperparameter combination here, and have provided figures of performance gains for all hyperparameters in Appendix~\ref*{app:hpsearch}. It can be clearly observed that SSL-\ours outperforms the base model by large margins across most datasets and for both SimSiam and SimCLR. On \ds{cmnist}, our performance is very close to the baseline model and we do not see significant improvement. We hypothesize that this is due to the fact that the base encoder on the easier \ds{cmnist} dataset is already quite performant. On datasets where the base encoder performs poorly such as \ds{metashift} and \ds{spurcifar10}, our approach improves the performance by at least 10\% over base SimSiam. On a dataset of a larger scale like \ds{celebA}, \ours still improves upon a strong encoder baseline.

\begin{table*}[t!]

\centering
\caption{Worst-group accuracy (\%) of SSL-Base and \ours for SimSiam and SimCLR pre-training. Results for average accuracy can be found in Table~\ref{tab:prune-erm-gap}.}
\label{tab:prune-all}
\adjustbox{max width=0.8\textwidth}{%
\begin{tabular}{@{}lrrrr@{}}
\toprule
              & \multicolumn{2}{c}{\textbf{SimSiam}} & \multicolumn{2}{c}{\textbf{SimCLR}} \\ \cmidrule{2-3} \cmidrule{4-5}
              & \textsc{SSL-base}      & \textsc{SSL-Late-TVG}       & \textsc{SSL-base}        & \textsc{SSL-Late-TVG}     \\ \cmidrule(l){2-3} \cmidrule(l){4-5}
\ds{celebA}       & 77.5                               & \textbf{83.1}                                   & 76.7                                                   & \textbf{82.2}                                   \\
\ds{cmnist}       & 80.7                               & \textbf{83.1}                                   & 81.7                                                   & \textbf{83.8}                                   \\
\ds{metashift}    & 42.3                               & \textbf{79.6}                                   & 45.5                                                   & \textbf{59.3}                                   \\
\ds{spurcifar10} & 43.4                               & \textbf{61.4}                                   & 36.5                                                   & \textbf{40.4}                                  \\
\ds{waterbirds}   & 48.3                               & \textbf{56.3}                                   & 43.8                                                   & \textbf{55.4}   \\ \bottomrule
\end{tabular}
}
\end{table*}

Further, we find that \ours closes the gap in performance to supervised pretraining (Table ~\ref{tab:prune-erm-gap}). We emphasize that this is an unfair comparison to begin with, since supervised pretraining requires labeled data whereas SSL does not, hence reducing the annotation budget drastically. Regardless, we find that \ours narrows the gap between the SSL baseline and the ERM model significantly -- 17\% relative improvement for \ds{cmnist} to 50\% in the case of \ds{spurcifar10}. In the case of \ds{celebA}, we even outperform the ERM baseline.

\subsubsection{SSL downstream linear performance is less reliant on a balanced downstream dataset}

Traditional approaches that mitigate spurious correlations in ERM-based settings assume that the downstream training set is balanced~\citep{kirichenko2022last}. However, this still requires knowledge of the spurious feature, which we may not always have in practice. 
In this experiment, we challenge this assumption and analyze how SSL models behave when the downstream training set is imbalanced.

We vary the proportion of minority groups in the downstream training set, by first downsampling the training set to have the same number of samples across groups, and second randomly sampling minority groups with weight $\lambda$ (x-axis in Figure~\ref{fig:minorchange}) and majority groups with weights $1-\lambda$. We measure the worst group accuracy of the trained linear models for each dataset. We show the results on \ds{metashift} in Figure~\ref{fig:minorchange}, comparing the performance of SSL-Base and SSL-\ours. We can observe that \ours outperforms the baseline across a range of minority weights -- implying that \ours is more robust to imbalances in downstream training data. This is a crucial aspect where \ours differs from other approaches in the supervised pretraining literature, such as DFR~\citep{kirichenko2022last}, which requires a balanced training set for the reweighting strategy to be successful. Similar results for other datasets and linear models are provided in in Appendix~\ref*{appdx:minority}.

\subsubsection{\ours reduces Spurious Connectivity in the Representation Space}
\label{app:spconres}

Finally, we relate our method back to the theoretical analysis presented in Section \ref{sec:spurious_downstream_failures}, by computing the connectivity of the \textit{representation space} learned by the SSL models, using the procedure outlined in Section \ref{app:spconexpt}. In Table \ref{tab:connectivity}, we find that \ours empirically reduces the spurious connectivity, while increasing the invariant connectivity, for all datasets. Thus, we have shown that \ours successfully augments the representation space to induce desired invariances.

\begin{table}[htbp]
\caption{We report the error of classifiers trained to distinguish between the \emph{representations} of two subgroups as a proxy for connectivity terms. We find that \ours decreases spurious connectivity while increasing invariant connectivity in comparison to the baseline.}
\label{tab:connectivity}
\resizebox{0.7\linewidth}{!}{
\begin{tabular}{@{}l|l|ccc@{}}
\toprule
\textbf{Dataset} & \textbf{Representation Space}   & \multicolumn{1}{c}{\textbf{\begin{tabular}[c]{@{}c@{}}Spurious \\ Connectivity\end{tabular}}} & \multicolumn{1}{c}{\textbf{\begin{tabular}[c]{@{}c@{}}Invariant \\ Connectivity\end{tabular}}} & \multicolumn{1}{c}{\textbf{\begin{tabular}[c]{@{}c@{}}Opposite \\ Connectivity\end{tabular}}} \\ \midrule
\ds{celebA} &  \textsc{SSL-Base}    & 18.9                  & 15.7                   & 8.3 \\
&  \textsc{SSL-Late-TVG}  & 15.8                  & 17.9                   & 8.0 \\ \midrule
\ds{cmnist}  &  \textsc{SSL-Base}  & 37.3                  & 3.2                    & 2.7 \\ 
&  \textsc{SSL-Late-TVG}  & 34.8                  & 3.8                    & 3.0 \\ \midrule
\ds{metashift} &  \textsc{SSL-Base} & 28.6                  & 21.4                   & 21.8 \\ 
 &	\textsc{SSL-Late-TVG} & 27.3                  & 27.3                   & 21.3 \\ \midrule
\ds{waterbirds} &  \textsc{SSL-Base} & 44.9                  & 9.4                   & 8.4 \\ 
& \textsc{SSL-Late-TVG} & 44.6                  & 13.5                   & 12.8 \\ \bottomrule
\end{tabular}
}
\centering
\end{table}

\section{Conclusion}
In this paper, we have investigated the impact of spurious correlations on self-supervised learning (SSL) pre-training and proposed a new approach, called \ours to address the issue. Our experiments demonstrated that spurious correlations caused by data augmentation can lead to spurious connectivity and hinder the model's ability to learn core features, which ultimately impacts downstream task performance. We have shown that traditional debiasing techniques, such as re-sampling, are not effective in mitigating the impact of spurious correlations in SSL pre-training. In contrast, \ours effectively improves the worst-group performance in downstream tasks by inducing invariance to spurious features in the representation space throughout training. Our approach does not require access to group or label information during training and can be applied to large-scale, imbalanced datasets with spurious correlations. We believe our work will help advance the field of SSL pre-training and encourage future research in developing methods that are robust to spurious correlations.

\clearpage
\bibliography{iclr2024_conference}

\begin{thebibliography}{73}
\providecommand{\natexlab}[1]{#1}
\providecommand{\url}[1]{\texttt{#1}}
\expandafter\ifx\csname urlstyle\endcsname\relax
  \providecommand{\doi}[1]{doi: #1}\else
  \providecommand{\doi}{doi: \begingroup \urlstyle{rm}\Url}\fi

\bibitem[Abnar et~al.(2021)Abnar, Dehghani, Neyshabur, and Sedghi]{abnar2021exploring}
Samira Abnar, Mostafa Dehghani, Behnam Neyshabur, and Hanie Sedghi.
\newblock Exploring the limits of large scale pre-training.
\newblock \emph{arXiv preprint arXiv:2110.02095}, 2021.

\bibitem[Agarwal et~al.(2021)Agarwal, Krueger, Clark, Radford, Kim, and Brundage]{agarwal2021evaluating}
Sandhini Agarwal, Gretchen Krueger, Jack Clark, Alec Radford, Jong~Wook Kim, and Miles Brundage.
\newblock Evaluating clip: towards characterization of broader capabilities and downstream implications.
\newblock \emph{arXiv preprint arXiv:2108.02818}, 2021.

\bibitem[Arjovsky et~al.(2019)Arjovsky, Bottou, Gulrajani, and Lopez-Paz]{arjovsky2019invariant}
Martin Arjovsky, L{\'e}on Bottou, Ishaan Gulrajani, and David Lopez-Paz.
\newblock Invariant risk minimization.
\newblock \emph{arXiv preprint arXiv:1907.02893}, 2019.

\bibitem[Bordes et~al.(2023)Bordes, Balestriero, Garrido, Bardes, and Vincent]{bordes2023guillotine}
Florian Bordes, Randall Balestriero, Quentin Garrido, Adrien Bardes, and Pascal Vincent.
\newblock Guillotine regularization: Why removing layers is needed to improve generalization in self-supervised learning.
\newblock \emph{Transactions on Machine Learning Research}, 2023.

\bibitem[Calude \& Longo(2017)Calude and Longo]{calude2017deluge}
Cristian~S Calude and Giuseppe Longo.
\newblock The deluge of spurious correlations in big data.
\newblock \emph{Foundations of science}, 22:\penalty0 595--612, 2017.

\bibitem[Caron et~al.(2020)Caron, Misra, Mairal, Goyal, Bojanowski, and Joulin]{swav}
Mathilde Caron, Ishan Misra, Julien Mairal, Priya Goyal, Piotr Bojanowski, and Armand Joulin.
\newblock Unsupervised learning of visual features by contrasting cluster assignments.
\newblock \emph{arXiv preprint arXiv:2006.09882}, 2020.

\bibitem[Caron et~al.(2021)Caron, Touvron, Misra, J{\'e}gou, Mairal, Bojanowski, and Joulin]{dino}
Mathilde Caron, Hugo Touvron, Ishan Misra, Herv{\'e} J{\'e}gou, Julien Mairal, Piotr Bojanowski, and Armand Joulin.
\newblock Emerging properties in self-supervised vision transformers.
\newblock In \emph{Proceedings of the IEEE/CVF international conference on computer vision}, pp.\  9650--9660, 2021.

\bibitem[Chen et~al.(2020{\natexlab{a}})Chen, Kornblith, Norouzi, and Hinton]{chen2020simple}
Ting Chen, Simon Kornblith, Mohammad Norouzi, and Geoffrey Hinton.
\newblock A simple framework for contrastive learning of visual representations.
\newblock In \emph{International conference on machine learning}, pp.\  1597--1607. PMLR, 2020{\natexlab{a}}.

\bibitem[Chen et~al.(2021)Chen, Luo, and Li]{chen2021intriguing}
Ting Chen, Calvin Luo, and Lala Li.
\newblock Intriguing properties of contrastive losses.
\newblock \emph{Advances in Neural Information Processing Systems}, 34:\penalty0 11834--11845, 2021.

\bibitem[{Chen} \& {He}(2020){Chen} and {He}]{simsiam}
Xinlei {Chen} and Kaiming {He}.
\newblock {Exploring Simple Siamese Representation Learning}.
\newblock \emph{arXiv e-prints}, art. arXiv:2011.10566, November 2020.

\bibitem[Chen et~al.(2020{\natexlab{b}})Chen, Fan, Girshick, and He]{chen2020improved}
Xinlei Chen, Haoqi Fan, Ross Girshick, and Kaiming He.
\newblock Improved baselines with momentum contrastive learning.
\newblock \emph{arXiv preprint arXiv:2003.04297}, 2020{\natexlab{b}}.

\bibitem[Creager et~al.(2021)Creager, Jacobsen, and Zemel]{creager2021environment}
Elliot Creager, J{\"o}rn-Henrik Jacobsen, and Richard Zemel.
\newblock Environment inference for invariant learning.
\newblock In \emph{International Conference on Machine Learning}, pp.\  2189--2200. PMLR, 2021.

\bibitem[DeGrave et~al.(2021)DeGrave, Janizek, and Lee]{degrave2021ai}
Alex~J DeGrave, Joseph~D Janizek, and Su-In Lee.
\newblock Ai for radiographic covid-19 detection selects shortcuts over signal.
\newblock \emph{Nature Machine Intelligence}, 3\penalty0 (7):\penalty0 610--619, 2021.

\bibitem[Deng et~al.(2009)Deng, Dong, Socher, Li, Li, and Fei-Fei]{deng2009imagenet}
Jia Deng, Wei Dong, Richard Socher, Li-Jia Li, Kai Li, and Li~Fei-Fei.
\newblock Imagenet: A large-scale hierarchical image database.
\newblock In \emph{2009 IEEE conference on computer vision and pattern recognition}, pp.\  248--255. Ieee, 2009.

\bibitem[Doersch et~al.(2015)Doersch, Gupta, and Efros]{doersch2015unsupervised}
Carl Doersch, Abhinav Gupta, and Alexei~A Efros.
\newblock Unsupervised visual representation learning by context prediction.
\newblock In \emph{Proceedings of the IEEE international conference on computer vision}, pp.\  1422--1430, 2015.

\bibitem[Duchi et~al.(2019)Duchi, Hashimoto, and Namkoong]{duchi2019distributionally}
John~C Duchi, Tatsunori Hashimoto, and Hongseok Namkoong.
\newblock Distributionally robust losses against mixture covariate shifts.
\newblock \emph{Under review}, 2, 2019.

\bibitem[Fan et~al.(2014)Fan, Han, and Liu]{fan2014challenges}
Jianqing Fan, Fang Han, and Han Liu.
\newblock Challenges of big data analysis.
\newblock \emph{National science review}, 1\penalty0 (2):\penalty0 293--314, 2014.

\bibitem[Gao et~al.(2023)Gao, Sagawa, Koh, Hashimoto, and Liang]{gao2023out}
Irena Gao, Shiori Sagawa, Pang~Wei Koh, Tatsunori Hashimoto, and Percy Liang.
\newblock Out-of-domain robustness via targeted augmentations.
\newblock \emph{arXiv preprint arXiv:2302.11861}, 2023.

\bibitem[Geirhos et~al.(2018)Geirhos, Rubisch, Michaelis, Bethge, Wichmann, and Brendel]{geirhos2018imagenet}
Robert Geirhos, Patricia Rubisch, Claudio Michaelis, Matthias Bethge, Felix~A Wichmann, and Wieland Brendel.
\newblock Imagenet-trained cnns are biased towards texture; increasing shape bias improves accuracy and robustness.
\newblock \emph{arXiv preprint arXiv:1811.12231}, 2018.

\bibitem[{Grill} et~al.(2020){Grill}, {Strub}, {Altch{\'e}}, {Tallec}, {Richemond}, {Buchatskaya}, {Doersch}, {Avila Pires}, {Guo}, {Gheshlaghi Azar}, {Piot}, {Kavukcuoglu}, {Munos}, and {Valko}]{byol}
Jean-Bastien {Grill}, Florian {Strub}, Florent {Altch{\'e}}, Corentin {Tallec}, Pierre~H. {Richemond}, Elena {Buchatskaya}, Carl {Doersch}, Bernardo {Avila Pires}, Zhaohan~Daniel {Guo}, Mohammad {Gheshlaghi Azar}, Bilal {Piot}, Koray {Kavukcuoglu}, R{\'e}mi {Munos}, and Michal {Valko}.
\newblock {Bootstrap your own latent: A new approach to self-supervised Learning}.
\newblock \emph{arXiv e-prints}, art. arXiv:2006.07733, June 2020.

\bibitem[Gururangan et~al.(2018)Gururangan, Swayamdipta, Levy, Schwartz, Bowman, and Smith]{gururangan2018annotation}
Suchin Gururangan, Swabha Swayamdipta, Omer Levy, Roy Schwartz, Samuel~R Bowman, and Noah~A Smith.
\newblock Annotation artifacts in natural language inference data.
\newblock \emph{arXiv preprint arXiv:1803.02324}, 2018.

\bibitem[HaoChen et~al.(2021)HaoChen, Wei, Gaidon, and Ma]{haochen2021provable}
Jeff~Z HaoChen, Colin Wei, Adrien Gaidon, and Tengyu Ma.
\newblock Provable guarantees for self-supervised deep learning with spectral contrastive loss.
\newblock \emph{Advances in Neural Information Processing Systems}, 34:\penalty0 5000--5011, 2021.

\bibitem[Hashimoto et~al.(2018)Hashimoto, Srivastava, Namkoong, and Liang]{hashimoto2018fairness}
Tatsunori Hashimoto, Megha Srivastava, Hongseok Namkoong, and Percy Liang.
\newblock Fairness without demographics in repeated loss minimization.
\newblock In \emph{International Conference on Machine Learning}, pp.\  1929--1938. PMLR, 2018.

\bibitem[He et~al.(2019)He, Fan, Wu, Xie, and Girshick]{moco}
Kaiming He, Haoqi Fan, Yuxin Wu, Saining Xie, and Ross Girshick.
\newblock Momentum contrast for unsupervised visual representation learning.
\newblock \emph{arXiv:1911.05722}, 2019.

\bibitem[Hooker et~al.(2019)Hooker, Courville, Clark, Dauphin, and Frome]{hooker2019compressed}
Sara Hooker, Aaron Courville, Gregory Clark, Yann Dauphin, and Andrea Frome.
\newblock What do compressed deep neural networks forget?
\newblock \emph{arXiv preprint arXiv:1911.05248}, 2019.

\bibitem[Idrissi et~al.(2021)Idrissi, Arjovsky, Pezeshki, and Lopez-Paz]{idrissi2021simple}
Badr~Youbi Idrissi, Martin Arjovsky, Mohammad Pezeshki, and David Lopez-Paz.
\newblock Simple data balancing achieves competitive worst-group-accuracy.
\newblock \emph{arXiv preprint arXiv:2110.14503}, 2021.

\bibitem[Jaiswal et~al.(2020)Jaiswal, Babu, Zadeh, Banerjee, and Makedon]{jaiswal2020survey}
Ashish Jaiswal, Ashwin~Ramesh Babu, Mohammad~Zaki Zadeh, Debapriya Banerjee, and Fillia Makedon.
\newblock A survey on contrastive self-supervised learning.
\newblock \emph{Technologies}, 9\penalty0 (1):\penalty0 2, 2020.

\bibitem[Jiang et~al.(2021{\natexlab{a}})Jiang, Chen, Chen, and Wang]{jiang2021improving}
Ziyu Jiang, Tianlong Chen, Ting Chen, and Zhangyang Wang.
\newblock Improving contrastive learning on imbalanced data via open-world sampling.
\newblock \emph{Advances in Neural Information Processing Systems}, 34, 2021{\natexlab{a}}.

\bibitem[Jiang et~al.(2021{\natexlab{b}})Jiang, Chen, Mortazavi, and Wang]{jiang2021self}
Ziyu Jiang, Tianlong Chen, Bobak~J Mortazavi, and Zhangyang Wang.
\newblock Self-damaging contrastive learning.
\newblock In \emph{International Conference on Machine Learning}, pp.\  4927--4939. PMLR, 2021{\natexlab{b}}.

\bibitem[Jing \& Tian(2020)Jing and Tian]{jing2020self}
Longlong Jing and Yingli Tian.
\newblock Self-supervised visual feature learning with deep neural networks: A survey.
\newblock \emph{IEEE transactions on pattern analysis and machine intelligence}, 43\penalty0 (11):\penalty0 4037--4058, 2020.

\bibitem[Joshi et~al.(2023)Joshi, Yang, Xue, Yang, and Mirzasoleiman]{joshi2023towards}
Siddharth Joshi, Yu~Yang, Yihao Xue, Wenhan Yang, and Baharan Mirzasoleiman.
\newblock Towards mitigating spurious correlations in the wild: A benchmark \& a more realistic dataset.
\newblock \emph{arXiv preprint arXiv:2306.11957}, 2023.

\bibitem[Kirichenko et~al.(2022)Kirichenko, Izmailov, and Wilson]{kirichenko2022last}
Polina Kirichenko, Pavel Izmailov, and Andrew~Gordon Wilson.
\newblock Last layer re-training is sufficient for robustness to spurious correlations.
\newblock \emph{arXiv preprint arXiv:2204.02937}, 2022.

\bibitem[Koh et~al.(2021)Koh, Sagawa, Marklund, Xie, Zhang, Balsubramani, Hu, Yasunaga, Phillips, Gao, et~al.]{koh2021wilds}
Pang~Wei Koh, Shiori Sagawa, Henrik Marklund, Sang~Michael Xie, Marvin Zhang, Akshay Balsubramani, Weihua Hu, Michihiro Yasunaga, Richard~Lanas Phillips, Irena Gao, et~al.
\newblock Wilds: A benchmark of in-the-wild distribution shifts.
\newblock In \emph{International Conference on Machine Learning}, pp.\  5637--5664. PMLR, 2021.

\bibitem[Lahoti et~al.(2020)Lahoti, Beutel, Chen, Lee, Prost, Thain, Wang, and Chi]{lahoti2020fairness}
Preethi Lahoti, Alex Beutel, Jilin Chen, Kang Lee, Flavien Prost, Nithum Thain, Xuezhi Wang, and Ed~H. Chi.
\newblock Fairness without demographics through adversarially reweighted learning, 2020.

\bibitem[Lee et~al.(2022{\natexlab{a}})Lee, Chen, Tajwar, Kumar, Yao, Liang, and Finn]{lee2022surgical}
Yoonho Lee, Annie~S Chen, Fahim Tajwar, Ananya Kumar, Huaxiu Yao, Percy Liang, and Chelsea Finn.
\newblock Surgical fine-tuning improves adaptation to distribution shifts.
\newblock \emph{arXiv preprint arXiv:2210.11466}, 2022{\natexlab{a}}.

\bibitem[Lee et~al.(2022{\natexlab{b}})Lee, Yao, and Finn]{lee2022diversify}
Yoonho Lee, Huaxiu Yao, and Chelsea Finn.
\newblock Diversify and disambiguate: Learning from underspecified data.
\newblock \emph{arXiv preprint arXiv:2202.03418}, 2022{\natexlab{b}}.

\bibitem[Liang \& Zou(2022)Liang and Zou]{liang2022metashift}
Weixin Liang and James Zou.
\newblock Metashift: A dataset of datasets for evaluating contextual distribution shifts and training conflicts.
\newblock \emph{arXiv preprint arXiv:2202.06523}, 2022.

\bibitem[Liu et~al.(2021{\natexlab{a}})Liu, Haghgoo, Chen, Raghunathan, Koh, Sagawa, Liang, and Finn]{liu2021just}
Evan~Z Liu, Behzad Haghgoo, Annie~S Chen, Aditi Raghunathan, Pang~Wei Koh, Shiori Sagawa, Percy Liang, and Chelsea Finn.
\newblock Just train twice: Improving group robustness without training group information.
\newblock In \emph{International Conference on Machine Learning}, pp.\  6781--6792. PMLR, 2021{\natexlab{a}}.

\bibitem[Liu et~al.(2021{\natexlab{b}})Liu, HaoChen, Gaidon, and Ma]{liu2021self}
Hong Liu, Jeff~Z HaoChen, Adrien Gaidon, and Tengyu Ma.
\newblock Self-supervised learning is more robust to dataset imbalance.
\newblock \emph{arXiv preprint arXiv:2110.05025}, 2021{\natexlab{b}}.

\bibitem[Liu et~al.(2015)Liu, Luo, Wang, and Tang]{liu2015deep}
Ziwei Liu, Ping Luo, Xiaogang Wang, and Xiaoou Tang.
\newblock Deep learning face attributes in the wild.
\newblock In \emph{Proceedings of the IEEE international conference on computer vision}, pp.\  3730--3738, 2015.

\bibitem[McCoy et~al.(2019)McCoy, Pavlick, and Linzen]{mccoy2019right}
R~Thomas McCoy, Ellie Pavlick, and Tal Linzen.
\newblock Right for the wrong reasons: Diagnosing syntactic heuristics in natural language inference.
\newblock \emph{arXiv preprint arXiv:1902.01007}, 2019.

\bibitem[Meehan et~al.(2023)Meehan, Bordes, Vincent, Chaudhuri, and Guo]{meehan2023ssl}
Casey Meehan, Florian Bordes, Pascal Vincent, Kamalika Chaudhuri, and Chuan Guo.
\newblock Do ssl models have d\'ej\`a vu? a case of unintended memorization in self-supervised learning, 2023.

\bibitem[Menon et~al.(2021)Menon, Rawat, and Kumar]{menon2021overparameterisation}
Aditya~Krishna Menon, Ankit~Singh Rawat, and Sanjiv Kumar.
\newblock Overparameterisation and worst-case generalisation: friend or foe?
\newblock In \emph{International Conference on Learning Representations}, 2021.
\newblock URL \url{https://openreview.net/forum?id=jphnJNOwe36}.

\bibitem[Moayeri et~al.(2022)Moayeri, Singla, and Feizi]{moayeri2022hard}
Mazda Moayeri, Sahil Singla, and Soheil Feizi.
\newblock Hard imagenet: Segmentations for objects with strong spurious cues.
\newblock \emph{Advances in Neural Information Processing Systems}, 35:\penalty0 10068--10077, 2022.

\bibitem[Nagarajan et~al.(2020)Nagarajan, Andreassen, and Neyshabur]{nagarajan2020understanding}
Vaishnavh Nagarajan, Anders Andreassen, and Behnam Neyshabur.
\newblock Understanding the failure modes of out-of-distribution generalization.
\newblock \emph{arXiv preprint arXiv:2010.15775}, 2020.

\bibitem[Nam et~al.(2020)Nam, Cha, Ahn, Lee, and Shin]{nam2020learning}
Junhyun Nam, Hyuntak Cha, Sungsoo Ahn, Jaeho Lee, and Jinwoo Shin.
\newblock Learning from failure: De-biasing classifier from biased classifier.
\newblock \emph{Advances in Neural Information Processing Systems}, 33:\penalty0 20673--20684, 2020.

\bibitem[Oord et~al.(2018)Oord, Li, and Vinyals]{oord2018representation}
Aaron van~den Oord, Yazhe Li, and Oriol Vinyals.
\newblock Representation learning with contrastive predictive coding.
\newblock \emph{arXiv preprint arXiv:1807.03748}, 2018.

\bibitem[Oquab et~al.(2023)Oquab, Darcet, Moutakanni, Vo, Szafraniec, Khalidov, Fernandez, Haziza, Massa, El-Nouby, et~al.]{oquab2023dinov2}
Maxime Oquab, Timoth{\'e}e Darcet, Th{\'e}o Moutakanni, Huy Vo, Marc Szafraniec, Vasil Khalidov, Pierre Fernandez, Daniel Haziza, Francisco Massa, Alaaeldin El-Nouby, et~al.
\newblock Dinov2: Learning robust visual features without supervision.
\newblock \emph{arXiv preprint arXiv:2304.07193}, 2023.

\bibitem[Robinson et~al.(2021)Robinson, Sun, Yu, Batmanghelich, Jegelka, and Sra]{robinson2021can}
Joshua Robinson, Li~Sun, Ke~Yu, Kayhan Batmanghelich, Stefanie Jegelka, and Suvrit Sra.
\newblock Can contrastive learning avoid shortcut solutions?
\newblock \emph{arXiv preprint arXiv:2106.11230}, 2021.

\bibitem[Rosenfeld et~al.(2022)Rosenfeld, Ravikumar, and Risteski]{rosenfeld2022domain}
Elan Rosenfeld, Pradeep Ravikumar, and Andrej Risteski.
\newblock Domain-adjusted regression or: Erm may already learn features sufficient for out-of-distribution generalization.
\newblock \emph{arXiv preprint arXiv:2202.06856}, 2022.

\bibitem[Sagawa et~al.(2020{\natexlab{a}})Sagawa, Koh, Hashimoto, and Liang]{sagawa2020groupdro}
Shiori Sagawa, Pang~Wei Koh, Tatsunori~B. Hashimoto, and Percy Liang.
\newblock Distributionally robust neural networks for group shifts: On the importance of regularization for worst-case generalization, 2020{\natexlab{a}}.

\bibitem[Sagawa et~al.(2020{\natexlab{b}})Sagawa, Raghunathan, Koh, and Liang]{sagawa2020investigation}
Shiori Sagawa, Aditi Raghunathan, Pang~Wei Koh, and Percy Liang.
\newblock An investigation of why overparameterization exacerbates spurious correlations.
\newblock In \emph{International Conference on Machine Learning}, pp.\  8346--8356. PMLR, 2020{\natexlab{b}}.

\bibitem[Salman et~al.(2022)Salman, Jain, Ilyas, Engstrom, Wong, and Madry]{salman2022does}
Hadi Salman, Saachi Jain, Andrew Ilyas, Logan Engstrom, Eric Wong, and Aleksander Madry.
\newblock When does bias transfer in transfer learning?
\newblock \emph{arXiv preprint arXiv:2207.02842}, 2022.

\bibitem[Scalbert et~al.(2023)Scalbert, Vakalopoulou, and Couzini{\'e}-Devy]{scalbert2023improving}
Marin Scalbert, Maria Vakalopoulou, and Florent Couzini{\'e}-Devy.
\newblock Improving domain-invariance in self-supervised learning via batch styles standardization.
\newblock \emph{arXiv preprint arXiv:2303.06088}, 2023.

\bibitem[Selvaraju et~al.(2016)Selvaraju, Das, Vedantam, Cogswell, Parikh, and Batra]{selvaraju2016grad}
Ramprasaath~R Selvaraju, Abhishek Das, Ramakrishna Vedantam, Michael Cogswell, Devi Parikh, and Dhruv Batra.
\newblock Grad-cam: Why did you say that?
\newblock \emph{arXiv preprint arXiv:1611.07450}, 2016.

\bibitem[Shah et~al.(2020)Shah, Tamuly, Raghunathan, Jain, and Netrapalli]{shah2020pitfalls}
Harshay Shah, Kaustav Tamuly, Aditi Raghunathan, Prateek Jain, and Praneeth Netrapalli.
\newblock The pitfalls of simplicity bias in neural networks.
\newblock \emph{Advances in Neural Information Processing Systems}, 33:\penalty0 9573--9585, 2020.

\bibitem[Shen et~al.(2022)Shen, Jones, Kumar, Xie, HaoChen, Ma, and Liang]{shen2022connect}
Kendrick Shen, Robbie~M Jones, Ananya Kumar, Sang~Michael Xie, Jeff~Z HaoChen, Tengyu Ma, and Percy Liang.
\newblock Connect, not collapse: Explaining contrastive learning for unsupervised domain adaptation.
\newblock In \emph{International Conference on Machine Learning}, pp.\  19847--19878. PMLR, 2022.

\bibitem[Singla \& Feizi(2021)Singla and Feizi]{singla2021salient}
Sahil Singla and Soheil Feizi.
\newblock Salient imagenet: How to discover spurious features in deep learning?
\newblock \emph{arXiv preprint arXiv:2110.04301}, 2021.

\bibitem[Song et~al.(2019)Song, Kalluri, Grover, Zhao, and Ermon]{song2019learning}
Jiaming Song, Pratyusha Kalluri, Aditya Grover, Shengjia Zhao, and Stefano Ermon.
\newblock Learning controllable fair representations.
\newblock In \emph{The 22nd International Conference on Artificial Intelligence and Statistics}, pp.\  2164--2173. PMLR, 2019.

\bibitem[Tamkin et~al.(2021)Tamkin, Liu, Lu, Fein, Schultz, and Goodman]{tamkin2021dabs}
Alex Tamkin, Vincent Liu, Rongfei Lu, Daniel Fein, Colin Schultz, and Noah Goodman.
\newblock Dabs: A domain-agnostic benchmark for self-supervised learning.
\newblock \emph{arXiv preprint arXiv:2111.12062}, 2021.

\bibitem[Torralba \& Efros(2011)Torralba and Efros]{torralba2011unbiased}
Antonio Torralba and Alexei~A Efros.
\newblock Unbiased look at dataset bias.
\newblock In \emph{CVPR 2011}, pp.\  1521--1528. IEEE, 2011.

\bibitem[Tsai et~al.(2020)Tsai, Wu, Salakhutdinov, and Morency]{tsai2020demystifying}
Yao-Hung~Hubert Tsai, Yue Wu, Ruslan Salakhutdinov, and Louis-Philippe Morency.
\newblock Demystifying self-supervised learning: An information-theoretical framework.
\newblock \emph{arXiv e-prints}, pp.\  arXiv--2006, 2020.

\bibitem[Tu et~al.(2020)Tu, Lalwani, Gella, and He]{tu2020empirical}
Lifu Tu, Garima Lalwani, Spandana Gella, and He~He.
\newblock An empirical study on robustness to spurious correlations using pre-trained language models.
\newblock \emph{Transactions of the Association for Computational Linguistics}, 8:\penalty0 621--633, 2020.

\bibitem[Van~Horn et~al.(2021)Van~Horn, Cole, Beery, Wilber, Belongie, and Mac~Aodha]{van2021benchmarking}
Grant Van~Horn, Elijah Cole, Sara Beery, Kimberly Wilber, Serge Belongie, and Oisin Mac~Aodha.
\newblock Benchmarking representation learning for natural world image collections.
\newblock In \emph{Proceedings of the IEEE/CVF conference on computer vision and pattern recognition}, pp.\  12884--12893, 2021.

\bibitem[Wah et~al.(2011)Wah, Branson, Welinder, Perona, and Belongie]{wah2011caltech}
Catherine Wah, Steve Branson, Peter Welinder, Pietro Perona, and Serge Belongie.
\newblock The caltech-ucsd birds-200-2011 dataset.
\newblock 2011.

\bibitem[Wang et~al.(2021)Wang, Yue, Huang, Sun, and Zhang]{wang2021self}
Tan Wang, Zhongqi Yue, Jianqiang Huang, Qianru Sun, and Hanwang Zhang.
\newblock Self-supervised learning disentangled group representation as feature.
\newblock \emph{Advances in Neural Information Processing Systems}, 34:\penalty0 18225--18240, 2021.

\bibitem[Wang \& Culotta(2020)Wang and Culotta]{wang2020identifying}
Zhao Wang and Aron Culotta.
\newblock Identifying spurious correlations for robust text classification.
\newblock \emph{arXiv preprint arXiv:2010.02458}, 2020.

\bibitem[Yang et~al.(2023)Yang, Zhang, Katabi, and Ghassemi]{yang2023change}
Yuzhe Yang, Haoran Zhang, Dina Katabi, and Marzyeh Ghassemi.
\newblock Change is hard: A closer look at subpopulation shift.
\newblock In \emph{International Conference on Machine Learning}, 2023.

\bibitem[Zbontar et~al.(2021)Zbontar, Jing, Misra, LeCun, and Deny]{zbontar2021barlow}
Jure Zbontar, Li~Jing, Ishan Misra, Yann LeCun, and St{\'e}phane Deny.
\newblock Barlow twins: Self-supervised learning via redundancy reduction.
\newblock In \emph{International Conference on Machine Learning}, pp.\  12310--12320. PMLR, 2021.

\bibitem[Zech et~al.(2018)Zech, Badgeley, Liu, Costa, Titano, and Oermann]{zech2018variable}
John~R Zech, Marcus~A Badgeley, Manway Liu, Anthony~B Costa, Joseph~J Titano, and Eric~Karl Oermann.
\newblock Variable generalization performance of a deep learning model to detect pneumonia in chest radiographs: a cross-sectional study.
\newblock \emph{PLoS medicine}, 15\penalty0 (11):\penalty0 e1002683, 2018.

\bibitem[Zhang et~al.(2021)Zhang, Ahuja, Xu, Wang, and Courville]{zhang2021can}
Dinghuai Zhang, Kartik Ahuja, Yilun Xu, Yisen Wang, and Aaron Courville.
\newblock Can subnetwork structure be the key to out-of-distribution generalization?
\newblock In \emph{International Conference on Machine Learning}, pp.\  12356--12367. PMLR, 2021.

\bibitem[Zhang et~al.(2022)Zhang, Sohoni, Zhang, Finn, and R{\'e}]{zhang2022correct}
Michael Zhang, Nimit~S Sohoni, Hongyang~R Zhang, Chelsea Finn, and Christopher R{\'e}.
\newblock Correct-n-contrast: A contrastive approach for improving robustness to spurious correlations.
\newblock \emph{arXiv preprint arXiv:2203.01517}, 2022.

\bibitem[Zhou et~al.(2022)Zhou, Vani, Larochelle, and Courville]{zhou2022fortuitous}
Hattie Zhou, Ankit Vani, Hugo Larochelle, and Aaron Courville.
\newblock Fortuitous forgetting in connectionist networks.
\newblock \emph{arXiv preprint arXiv:2202.00155}, 2022.

\end{thebibliography}
\bibliographystyle{iclr2024_conference}


\newpage
\appendix
\section{Limitations}

In this work, we validated our method on several benchmark datasets containing spurious correlations from prior work \citep{sagawa2020groupdro,liang2022metashift,yang2023change}. However, we recognize that the scale of these datasets are small, relative to typical SSL training corpora (e.g. ImageNet \citep{deng2009imagenet}). As these large datasets do not contain annotations of spurious features, we are unable to evaluate our method in these settings. In addition, we primarily focus on SimSiam~\cite{simsiam} in our experiments, as it does not rely on large batch sizes and shows improved performance for smaller datasets. Moreover, we expect \ours to perform best in cases where Siamese encoders coupled with stop-gradient operation are used when learning the representations. 

\section{\ours Algorithm}
\label{app:alg}
We provide an algorithm representation of our proposed method \ours in Section~\ref{sec:late_tvg} as follows. 

\begin{algorithm}[h]
\caption{Self-supervised Learning with \ours}\label{alg:model_transformation}
\begin{algorithmic}[1]
    \State \textbf{Inputs:} Encoder $f$ parameterized by $\theta = \{W_1, \dots, W_n\}$, Projection head and predictor $g$, Augmentation module $\mathcal{T}$, Threshold $L$, Pruning rate $a$, Training epochs $N$.
    
    \State Initialize $\tilde{f}$ with $\theta$
    
    \vspace{0.1in}
    \ForAll {$i = 1 \rightarrow N$}
        
        \State \textbf{Stage 1}: Self-supervised Training
            \quad \ForAll  {$i = 1 \rightarrow N$} \quad
            \State Draw two random augmentations $t, t^{\prime} \sim \mathcal{T}$
            \State $x_1 = t(x), x_2=t^{\prime}(x) $ \Comment{Generate views $x_1, x_2$ from input $x$ using augmentation $t$}

            \State $v_1 = f(x_1) $ \Comment{Obtain encoded features from normal encoder $f$}
            
            \State $\tilde{v}_2 = \tilde{f}(x_2)$ \Comment{Obtain encoded features from transformed encoder $\tilde{f}$}

            \State  $\mathcal{L} = \text{Loss}(v_1, \tilde{v}_2; g)$ \Comment{Calculate contrastive loss given views $v_1$ and $\tilde{v}_2$}

            \State Update $f, g$ to minimize $\mathcal{L}$ \Comment{Update the encoder and other SSL parameters}

            \EndFor

        \vspace{0.1in}
        \State \textbf{Stage 2}: Model Transformation

         \State \quad 
        Compute the mask $M_{L, a} = \{M^l_L \odot \text{Top}_{a}(W_l) \mid  l \in [n]\}$ 
        
        \quad \quad where  $\text{Top}_{a}(W_l)_{i, j} = \mathbb{I}(\lvert {W_{l_{(i, j)}} \lvert } \text{ in top } a\% \text{ of } \theta) $

        \State \quad Update $\tilde{f}$ with parameters $\tilde{\theta} = M_{L, a} \odot \theta $ \Comment{Magnitude pruning of weights}
    \EndFor

    \State \textbf{Return} encoder $\tilde{f}$ 
\end{algorithmic}
\end{algorithm}

\section{Theoretical Analysis of Spurious Connectivity}
\label{app:theory}
\paragraph{Setup}
 We consider the pre-text task of learning representations from unlabeled population data $\adata$ consisting of unknown groups $\mathcal{G}$ which are not equally represented. 
For a given downstream task with labeled samples, we assume that each $x\in \adata$ belongs to one of $c = |\mathcal{Y}|$ classes, and let $y: \adata\rightarrow [c]$ denote the ground-truth labeling function. Let us define $a: \adata \rightarrow [m]$ as the deterministic attribute function creating groups (of potential different sizes) as $\mathcal{G} = \mathcal{Y} \times \mathcal{S}.$

\paragraph{Spectral Contrastive Learning}
In order to investigate why the invariant feature can be suppressed in contrastive learning, we consider the setting from~\citet{haochen2021provable} -- Spectral Contrastive learning, which achieves similar empirical results to other contrastive learning methods and is easier for theoretical analysis. 

Given the set of all natural data or data without any augmentation $\ndata$, we use $\aug{\bar{x}}$ to denote the distribution of augmentations of $\bar{x}\in \ndata$. For instance, when $\bar{x}$ represents an image, $\aug{\bar{x}}$ can be the distribution of common augmentations that includes Gaussian blur, color distortion and random cropping.

Let $\pndata$ be the population distribution over $\ndata$ from which we draw training data and test our final performance.  For any two augmented data points $x, x'\in \adata$, the weight between a pair $\wpair{x}{x'}$ is the marginal probability of generating the pair $x$ and $x'$ from a random data point $\bar{x}\sim \pndata$:
\begin{align*}
\wpair{x}{x'} = \Exp{\bar{x}\sim \pndata}\left[\augp{x}{\bar{x}} \augp{x'}{\bar{x}}\right]
\end{align*}

Define expansion between two sets similar to \cite{haochen2021provable} as below: 
$$\phi(S_1, S_2) = \frac{\sum_{x \in S_1 , x'\in S_2} w_{xx'}}{\sum_{x\in  S_1} w_x}$$ where $w_x = \sum_{x'\in \mathcal{S}} w_{xx'}$. 
We note that this is similar to our definition of connectivity, where we have assumed the marginal distribution over $x$ is uniform, or $w_x = \frac{1}{N}$.

\paragraph{Toy Setup}
\label{appx:proof}

Let the ground-truth labeling function  $y$ and the deterministic attribute function $a$, determine the subgroup $g = (y(x), a(x))$ of a given sample $x$. We suppose we have $n$ samples from each subgroup, and that labels and attributes take binary values\footnote{For an ease of notation and operations}.

Suppose that each edge in the augmentation graph is given by connectivity terms $\al$, $\be$, $\ro$, $\ga$  as below:
\begin{align*}
    \forall x, x' \in \adata: \quad P_+(x, x') &= \mathds{1} ( a(x) = a(x') , y(x) = y(x')) \ro \\
    &+ \mathds{1} ( a(x) \neq a(x') , y(x) = y(x')) \al \\
    &+ \mathds{1} ( a(x) = a(x') , y(x) \neq y(x'))\be  \\
    &+ \mathds{1} ( a(x) \neq a(x') , y(x) \neq y(x')) \ga
\end{align*}

We suppose that each edge in the augmentation graph is deterministically equal to one of the connectivity terms, and make the following assumptions: 

\begin{enumerate}
    \item $\al  > \ga , \be > \ga $ -- The probability that augmentation changes the spurious attribute only, or the class only is both greater than the probability that augmentation changes both attribute and class (at the same time). 
    \item  $ \ro > \al, \ro > \be $ -- The probability that augmentation that keeps both attribute and class is greater than the probability that it changes the spurious attribute only, or the class only is both higher than the probability that augmentation changes both domain and class (at the same time). 
    \item $\al > \be$ or Assumption~\ref{assumption:distortion} -- The probability that augmentation changes the spurious feature is higher than the probability of it changing the class, as observed in~\ref{tab:connectivity}.
\end{enumerate}

\subsection{Proof of Lemma~\ref{lemma:main}}

\begin{proof}
Let the $A \in \R^{4n \times 4n}$ be the adjacency matrix of the simplified augmentation graph. 
It is easy to show that $A$ is equivalent to adjacency matrix $\Bar{A}$ up to a rotation where:
\begin{align*}
\Bar{A}= & (\be-\ga) \cdot I_2 \otimes\left(\mathbf{1}_2 \mathbf{1}_2^{\top}\right) \otimes\left(\mathbf{1}_n \mathbf{1}_n^{\top}\right) \\
& +(\al-\ga) \cdot\left(\mathbf{1}_2 \mathbf{1}_2^{\top}\right) \otimes I_2 \otimes\left(\mathbf{1}_n \mathbf{1}_n^{\top}\right) \\
& +(\ro-\be-\al+\ga) \cdot I_4 \otimes\left(\mathbf{1}_n \mathbf{1}_n^{\top}\right) \\
& +\ga \cdot\left(\mathbf{1}_4 \mathbf{1}_4^{\top}\right) \otimes\left(\mathbf{1}_n \mathbf{1}_n^{\top}\right) 
\end{align*}
Where $\mathbf{1}_k$ is used to denote the all-one vector of dimension $k$ and let $\bar{\mathbf{1}}_k$ be the normalized version. 

For the case of $n=1$, it is easy to show that the matrix is reduced to an adjacency matrix of 4 nodes, each in one group, where the first two rows/columns correspond to samples with the same spurious attribute, and odd or even rows correspond to samples that are from the same class, based on the placements of $\al$ and $\be$ in the matrix. 

Let $F$ be an embedding matrix with $u_x$ on the $x$-th row which corresponds to the embeddings of sample $x$, and consider the matrix factorization based form of the spectral contrastive loss as below
$$
\min _{F \in \mathbb{R}^{N \times k}} \mathcal{L}_{\mathrm{mf}}(F):=\left\|\bar{A}-F F^{\top}\right\|_F^2
$$

It is enough to compute the eigenvectors of $\Bar{{A}},$ to obtain $F$.  
It is easy to compute the eigenvectors of $\Bar{{A}}$ similar to~\cite{shen2022connect}. The set of four sets of eigenvectors would be as below:
\begin{itemize}
    \item For eigenvalue $\lambda_1 = \ro + \be + \al + \ga $, \; the eigenvector is $\bar{\mathbf{1}}_2 \otimes \bar{\mathbf{1}}_2 \otimes \bar{\mathbf{1}}_n$. 
    \item For eigenvalue $\lambda_2 = \ro + \be-\al-\ga$ \; the eigenvectors are $[1 \;-1]^T \otimes \bar{\mathbf{1}}_2 \otimes \bar{\mathbf{1}}_n$. 
    \item For eigenvalue $\lambda_3 = \ro - \be + \al - \ga $\; the eigenvectors are $\bar{\mathbf{1}}_2 \otimes [1 \; -1]^T \otimes \bar{\mathbf{1}}_n$. 
    \item $\lambda_4 = \ro - \be - \al + \ga $ which is smaller than the first three eigenvalues, given the above assumptions. 
\end{itemize}

Thus $F$ would be a rank-3 matrix with columns equal to $\sqrt{\lambda_i}$ multiplied by each eigenvector. Given the case of $n=1$ explained above and by induction, it is easy to show that $\lambda_2$ corresponds to the spurious attribute subspace, and $\lambda_3$ corresponds to the class. Projecting samples in $\bar{A}$ with representations as rows of $F$, onto the spurious subspace suggests that the spurious feature takes two values $\{ - \sqrt{\lambda_2}, \sqrt{\lambda_2} \}$, and similarly, the invariant feature takes two values $\{ - \sqrt{\lambda_3}, \sqrt{\lambda_3} \}$ in the representation space learned by spectral contrastive loss. 
\end{proof}

Intuitively, this means that with higher spurious connectivity ---or higher weights on edges connecting images that only share the same spurious attribute--- spectral clustering will learn representations of the population data based on the spurious feature, rather than the invariant feature. 


\section{Data and Models}
\subsection{Datasets}
\label{app:datasets}
We make use of the following four image datasets:
\begin{itemize}
        \item \ds{celebA}~\citep{liu2015deep}: Gender (Male, Female) is spuriously correlated with Hair color (blond hair, not blond hair). 
    \item \texttt{waterbirds}~\citep{sagawa2020groupdro}: Background (land, water) is spuriously correlated with bird type (landbird, waterbird). 
    \item \texttt{cmnist} (Colored MNIST): The color of the digit on the images is spuriously correlated with the binary class based on the number. This is the same setup as ~\citet{arjovsky2019invariant}, except with no label flipping. 
    \item \texttt{spurcifar10} (Spurious CIFAR10)~\citep{nagarajan2020understanding}: The color of lines on the images spuriously correlated with the class. 
    \item \texttt{metashift}~\citep{liang2022metashift} We consider the Cats vs Dogs task where Background (indoor, outdoor) is spuriously correlated with pet type (cat, dog). 
\end{itemize}

Note that each data contains both labels (or core attribute) $y$, and spurious attribute $a$. We then use the group information $g = (y, a)$ to partition dataset splits into groups. 

\subsection{Methods and Hyperparameters}
\label{app:hpsearch}
We use SimSiam~\citep{simsiam} with ResNet encoders to train both base models and \ours. We select ResNet-18 models as the backbone for all datasets except for \ds{celebA}, which we use ResNet-50 models. 

For each dataset, we use the following set of hyperparameters for SimSiam training. 

\begin{table}[h!]
\begin{tabular}{lllll}
\toprule
\textbf{Dataset} & \textbf{Learning Rate} & \textbf{Batch Size} & \textbf{Weight Decay} & \textbf{Number of Epochs} \\ \midrule
\ds{celebA}          & 0.01                   & 128                 & 1e-4                  & 400                       \\
\ds{cmnist}           & 1e-3                   & 128                 & 1e-5                  & 1000                      \\
\ds{metashift}        & 0.05                   & 256                 & 0.001                 & 400         \\
\ds{spurcifar10}    & 0.02                   & 128                 & 5e-4                  & 800                       \\
\ds{waterbirds}       & 0.01                   & 64                  & 1e-3                  & 800                       \\ \bottomrule
\end{tabular}
\end{table}

The specific augmentations that we used for learning the representations, are exactly similar to the SimSiam~\cite{simsiam} paper but without color jitter. 

Note that the model architecture and parameters for \textsc{SSL-Base} and \textsc{SSL-Late-TVG} are exactly the same, but \textsc{SSL-Late-TVG} uses the pruning hyperparameters to prune the encoder during training. 

\paragraph{Computational Cost} The SSL-LateTVG model updates the same number of parameters as SSL-Base during training, with the forward pass keeping both the original and pruned encoder. The pruning operation is cost O(n) where n is the number of parameters. So any FLOPs used for the extra pruning mechanism will be very small compared to a single forward pass.

\subsection{The Role of Downstream Regularization}
We investigate the impact of regularization techniques during downstream Linear probing. Interestingly, we find that the presence and type of regularization has a notable effect on the accuracy of the worst-performing group, with improvements of approximately 10\% on the \ds{celebA} dataset and 7\% on the \ds{metashift} dataset. We hypothesize that the minority samples contribute more to the variance of the linear models, and the additional regularization helps penalize them, leading to a reduction in the variance of the downstream models.

\label{app:reg}
\begin{table}[h]
\caption{Accuracy (\%) of SimSiam models pretrained on each dataset with random initialization.}
\resizebox{.6\linewidth}{!}{
\begin{tabular}{@{}crrrrrr@{}}
\toprule
\multicolumn{1}{l}{} & \multicolumn{3}{c}{\textbf{Average}}                              & \multicolumn{3}{c}{\textbf{Worst-Group}}                          \\ \midrule
\multicolumn{1}{l}{} & \multicolumn{1}{r}{None} & \multicolumn{1}{r}{L1} & \multicolumn{1}{r}{L2} & \multicolumn{1}{r}{None} & \multicolumn{1}{r}{L1} & \multicolumn{1}{r}{L2} \\ \cmidrule(l){2-4} \cmidrule(l){5-7} 

\ds{celebA}      & 78.5                     & 81.9                   & \textbf{82.8}                   & 66.1                     &\textbf{77.5}                   & 76.7                   \\
\ds{cmnist}      & 80.5                     & 80.8                   & \textbf{82.7}                   & 76.3                     & 78.9                   & \textbf{81.3}                   \\
\ds{metashift}   & 54.2                     & \textbf{59.8}                   & 56.3                   & 41.8                     & \textbf{48.1}                   & 45.6                   \\
\ds{spurcifar10} & 69.3                     & 72.5                   & \textbf{73.4}                   & 41.1                     & 45.1                   & \textbf{49.0}                   \\
\ds{waterbirds}  & \textbf{52.0}                     & 51.2                   & 50.5                   & 47.4                     & \textbf{47.5}                   & 47.2                   \\ \bottomrule
\end{tabular}
}
\centering
\label{table:regularization}
\end{table}

\section{Measuring Spurious Connectivity in Augmentations}
\label{app:spconexpt}

In this section, we present our methodology for measuring spurious connectivity in augmentations. We conduct experiments on four datasets, and our goal is to quantify the extent to which samples within the training set are connected to each other through the spurious attribute, as opposed to the core feature. 

To estimate the average connectivity between two groups, denoted as $g_1$ and $g_2$, specified by class-attribute pairs $(y, a)$ and $(y', a')$, we follow the algorithm outlined below:

Initially, we label all training examples belonging to group $g_1$ or class $y$ and attribute $a$ as 0, and all training examples belonging to group $g_2$. Next, we train a classifier to distinguish between the two groups. The error of this classifier would be a proxy for ``the probability of augmented images being assigned to the other group'', or how close they are in the augmentation space. 
Instead of training a large classifier from scratch for each pair, we use CLIP's representations in Section \ref{sec:empasmpt}, and assume that it is extracting all necessary features for distinguishing between the two groups. In Section \ref{app:spconres}, we instead use the representations learned by each SSL model.

We train a linear model on these features to distinguish between each of the two groups. It is important to note that the augmentations used in our experiments are the classical augmentations commonly employed in SimSiam, excluding Gaussian blur. Subsequently, we create the test set following a similar process, where images are labeled based on their group or class-attribute pairs. The trained linear classifier is evaluated on this strongly augmented test set. The test error of the classifier serves as an estimate for the connectivity between the two pairs, providing insights into the degree of connectivity based on the spurious attribute.

By applying this methodology to all four datasets, we obtain results regarding the average spurious connectivity compared to the invariant connectivity. Table~\ref{tab:connectivity} summarizes the findings, revealing that, across all datasets, the average spurious connectivity is higher than the invariant connectivity. Furthermore, we validate that both these connectivity values are higher than the probability of simultaneously changing both the spurious attribute and the invariant attribute. These observations indicate that the samples within the training set are more likely to be connected to each other through the spurious attribute, rather than the core feature. This finding suggests a preference of the contrastive loss for alignment based on the spurious attribute rather than class alignment.

\newpage

\section{Additional Results for \ours}
\subsection{\ours reduces background reliance in Hard ImageNet}
\label{app:hardin}



We evaluate \ours on the Hard ImageNet dataset~\citep{moayeri2022hard}, which consists of 15 challenging ImageNet classes where models rely heavily on spurious correlations. The authors provide spuriousness rankings that enable creating a balanced subset.

In our experiments, we train the SSL model on the full Hard ImageNet train split, and train the linear classifier on the spurious-balanced subset. This tests the model's ability to learn representations without exploiting spurious cues.

We then evaluate the downstream classifier on four different dataset splits as below:

\begin{itemize}
    \item \textbf{None}: Original test split
    \item \textbf{Gray}: The object region is grayed out by replacing RGB values with the mean RGB value. This removes texture/color cues.
\item \textbf{Gray BBox}: The object region is removed by replacing it with the mean RGB value of the surrounding bounding box region. This ablates shape cues.

\item \textbf{Tile}: The object region is replaced by tiling the surrounding bounding box region. This also ablates shape cues.

\end{itemize}

A classifier relying on the spurious (i.e. non-object) features will achieve high performance in all evaluation splits. However, a classifier relying on the invariant features should perform decently on the original test split, but exhibit greatly reduced accuracy on the other splits. Thus, we desire high accuracy for the None split, and low accuracy for the other three splits.


Comparing the results to section 7 from~\citep{moayeri2022hard}, we find that the gap between None and other three splits is already large in SSL-base, and SSL-LateTVG is further decreasing the accuracy in the spurious datasets. This shows that the SSL-LateTVG encoder relies less on the spurious feature to predict the labels, which degrades the performance on splits that try remove the core feature.

We do not tune the hyperparameters in this experiment, but we find that for all sets of hyperparameters, SSL-LateTVG results in lower downstream accuracy on Gray, Gray BBox, and Tile splits as shown in Table~\ref{tab:hin}. 

\vspace{0.1in}

\begin{table}[h]
\centering
\adjustbox{max width=0.8\textwidth}{%
\begin{tabular}{l | l | rrrr}
\toprule
Algorithm                    & Pruning threshold, percentage & \multicolumn{1}{l}{\textbf{None} $\uparrow$} & \multicolumn{1}{l}{\textbf{Gray} $\downarrow$} & \multicolumn{1}{l}{\textbf{Gray BBox} $\downarrow$}& \multicolumn{1}{l}{\textbf{Tile} $\downarrow$} \\ \midrule
\textsc{SSL-Base}                     & -               & 79.5                     & 61.6                     & 53.5                          & 58.1                     \\ \midrule
\multirow{4}{*}{\textsc{SSL-LateTVG}} & 46, 0.5         & 78.0                     & 59.5                     & 51.1                          & 52.1                     \\
                             & 47, 0.5         & 76.7                     & 59.1                     & 49.6                          & 54.4                     \\
                             & 48, 0.8         & 73.9                     & 56.1                     & 48.0                          & 51.3                     \\
                             & 49, 0.8         & 68.4                     & 50.7                     & 42.4                          & 44.7                    \\
\bottomrule
\end{tabular}
}
\caption{We train SimSiam models with a ResNet-50 backbone on unlabeled data from Hard ImageNet containing spurious correlation, we then train the downstream linear classifier on a balanced subset, and evaluate the downstream model on splits containing spurious features -- \ours degrades the performance on these splits, without hyperparameter tuning}
\label{tab:hin}
\end{table}

\subsection{\ours closes the gap to supervised pre-training}
Self-supervised pretraining has shown a lot of promise in bridging the gap to supervised approaches in general representation learning. In this section, we explore whether this trend holds true for pre-training with data containing spurious correlations. To perform this analysis, we start with the same encoder model and vary only the pretraining strategy while fixing other aspects of the training, such hyperparameter selection and model selection. 

\begin{table*}[h]
\centering
\caption{Accuracy (\%) of SSL models pre-trained on each dataset versus features of a supervised model: Representations obtained from the supervised featurizer are more predictive of the core feature than SimCLR and SimSiam featurizers}
\label{tab:erm-ss-sc}
\adjustbox{max width=0.8\textwidth}{%
\begin{tabular}{@{}lrrrrrr@{}}
\toprule
              & \multicolumn{3}{c}{\textbf{Average Accuracy}}                                                   & \multicolumn{3}{c}{\textbf{Worst-Group Accuracy}}                                                             \\ \midrule
              & \textbf{SimCLR} & \textbf{SimSiam} & \textbf{Supervised} & \textbf{SimCLR} & \textbf{SimSiam} & \textbf{Supervised} \\ \cmidrule(l){2-4}  \cmidrule(l){5-7}  
\ds{celebA}       & 82.1                       & 81.9                        & \textbf{91.9}                        & 76.7                         & 77.5                          & \textbf{81.7}                                 \\
\ds{cmnist}       & 82.5                       & 82.1                        & \textbf{98.4}                        & 81.7                         & 80.7                          & \textbf{94.9}                                  \\
\ds{metashift}    & 55.1                       & 55.8                        & \textbf{89.8}                        & 45.5                         & 42.3                          & \textbf{83.5}                                  \\
\ds{spurcifar10} & 69.3                       & 75.1                        & \textbf{89.9}                        & 36.5                         & 43.4                          & \textbf{79.6}                                  \\
\ds{waterbirds}   & 47.5                       & 50.7                        & \textbf{67.9}                        & 43.8                         & \textbf{48.3}                          & 41.1                                  \\ \bottomrule
\end{tabular}
}
\end{table*}

We emphasize that this is an unfair comparison to begin with, since supervised pretraining requires labeled data whereas SSL does not, hence reducing the annotation budget drastically as shown in table~\ref{tab:erm-ss-sc}. However, the goal of this experiment to understand to what extent do SSL models and specifically \ours, compare with ERM based supervised pretraining strategies.

Table~\ref{tab:prune-erm-gap} shows the results of our experiment -- we have compared both average and worst group accuracies for the SSL-based and ERM-based encoders across all our evaluation datasets. In terms of worst group accuracy it is clear that \ours narrows the gap between the SSL baseline and the ERM model significantly -- 17\% relative improvement for \ds{cmnist} to 50\% in the case of \ds{spurcifar10}. In the case of \ds{celebA}, we even outperform the ERM baseline. Similar to previous experiments, the relative boost in performance from \ours is higher for cases where the base encoder is weaker, indicating the strength of our final layer augmentation in extracting useful signal relevant to the core features during pretraining. 
\begin{table*}[ht!]
\centering
\caption{\ours with SimSiam closes the gap between SSL baseline and supervised pre-training on worst group and average accuracy.}
\adjustbox{max width=\textwidth}{%
\label{tab:prune-erm-gap}
\begin{tabular}{@{}lrrrrrr@{}}
\toprule
              & \multicolumn{3}{c}{\textbf{Average Accuracy}} & \multicolumn{3}{c}{\textbf{Worst-group Accuracy}} \\ \cmidrule(l){2-7}  
              & \textsc{SSL-base}      & \textsc{SSL-Late-TVG}      & \textsc{Supervised}       & \textsc{SSL-base}        & \textsc{SSL-Late-TVG}       & \textsc{Supervised}        \\ \cmidrule(l){2-4} \cmidrule(l){5-7}
\ds{celebA}       & 81.9          & 88.9             & 91.9      & 77.5            & 83.1              & 81.7       \\
\ds{cmnist}       & 82.1          & 80.6             & 98.4      & 80.7            & 83.1              & 94.9       \\
\ds{metashift}    & 55.8          & 70.1              & 89.8      & 42.3            & 79.6              & 83.5       \\
\ds{spurcifar10} & 75.1          & 76.1             & 89.9      & 43.4            & 61.4              & 79.6       \\
\ds{waterbirds}   & 50.7          & 54.8             & 67.9      & 48.3            & 56.3              & 41.1       \\ \bottomrule
\end{tabular}
}
\end{table*}

\subsection{SSL-\ours outperforms baseline across hyper-parameter settings} \label{subsec:hyperparam}
\begin{figure}[!htbp]%
    \centering
    {\includegraphics[width=0.45\textwidth]{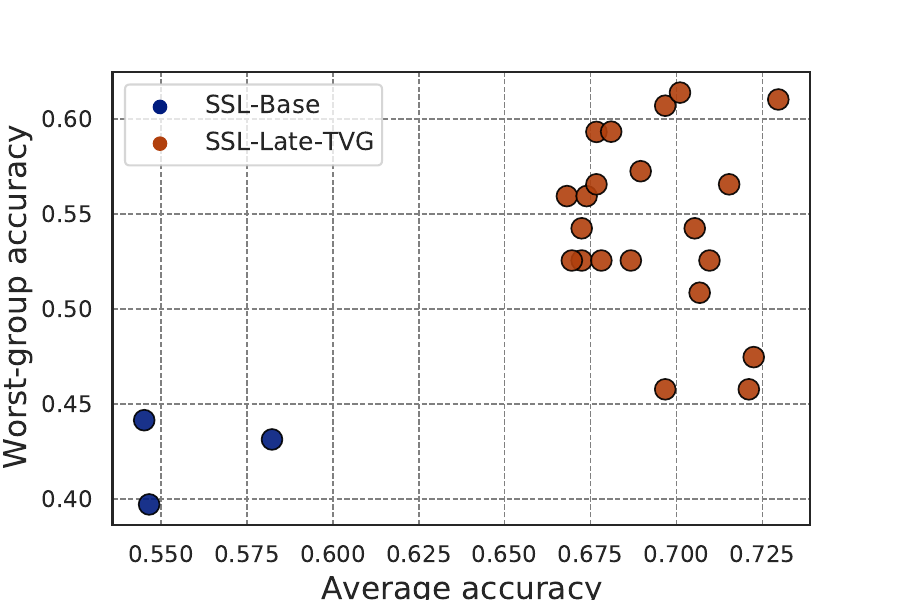}}
    \qquad
    \centering
    {\includegraphics[width=0.45\textwidth]{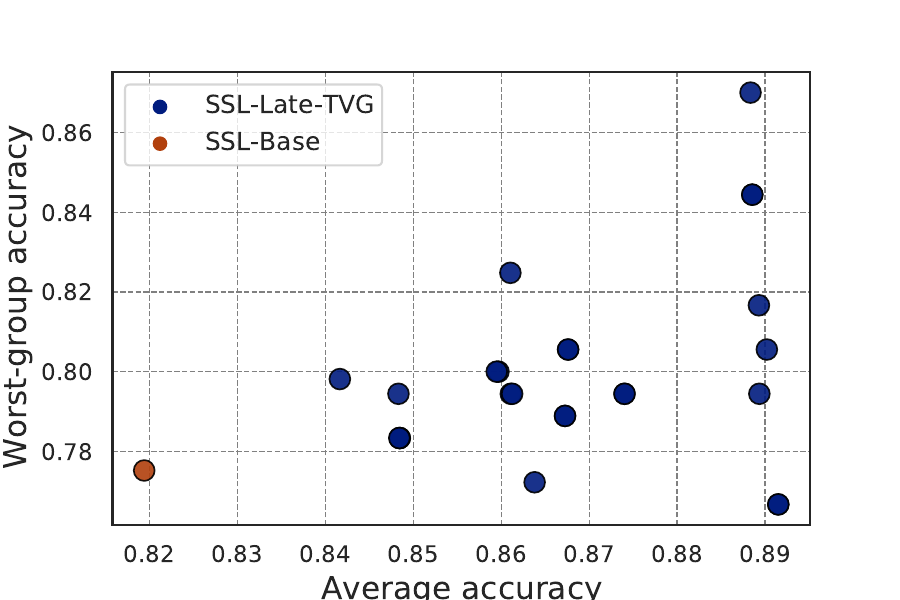}}
    \caption{Downstream worst-group accuracy of SSL-Late-TVG on the \ds{metashift} (left)  and \ds{celebA} (right) datasets as we vary the model pruning hyperparameters. }%
    \centering
    \label{fig:scatter}
\end{figure}

Disrupting the features and creating new views of the pairs is possible even with small amounts of pruning. We run a grid-search over the last three to five convolutional layers of ResNet models depending on the dataset, and choose pruning percentages varying between $[0.5, 0.7, 0.8, 0.9, 0.95]$. We find that in the \ds{metashift} dataset, \emph{all} hyperparameter settings improve the worst-group accuracy and outperform the baseline. Average and worst-group accuracies of different pruning hyperparameters on the \ds{metashift} and \ds{celebA} datasets is show in in figure~\ref{fig:scatter}.

Additionally, instead of choosing the best-performing model, we consider top 5 models across different pruning hyperparameters, and report the performance in Table~\ref{tab:main}. Even in this scenario, we observe large performance gains with \ours.

\begin{table*}[h!]
    \centering
    \caption{\ours improves baseline worst group and average accuracy of SSL models.}
\adjustbox{max width=0.85\textwidth}{%
    \begin{tabular}{@{}l|rrrrr@{}}
\toprule
\multicolumn{6}{c}{\textbf{Worst-group Accuracy}}                                                     \\ \midrule
                      & \ds{celebA}       & \ds{cmnist}       & \ds{metashift}    & 
                      \ds{spurcifar10} & \ds{waterbirds}   \\ \midrule
\textsc{SSL-Base}     & 77.52        & \textbf{80.7$\pm$2.71}  & 42.33$\pm$2.32 & 43.44$\pm$8.87 & 48.3$\pm$1.82  \\
\textsc{SSL-Late-TVG} & \textbf{81.83$\pm$1.75} & 77.18$\pm$1.59 & \textbf{60.34$\pm$0.97} & \textbf{54.58$\pm$1.74} & \textbf{51.87$\pm$2.37} \\ \midrule
\multicolumn{6}{c}{\textbf{Average Accuracy}}                                                             \\ \midrule
                      & \ds{celebA}       & \ds{cmnist}       & \ds{metashift}    & \ds{spurcifar10} & \ds{waterbirds}   \\ \midrule
\textsc{SSL-Base}     & 81.94        & \textbf{82.08$\pm$1.17} & 55.8$\pm$2.11  & 75.05$\pm$0.19 & 50.68$\pm$1.27 \\
\textsc{SSL-Late-TVG} & \textbf{87.32$\pm$1.46} & 79.74$\pm$1.19 & \textbf{69.7$\pm$2.09}  & \textbf{75.68$\pm$0.72} & \textbf{55.36$\pm$0.72} \\ \bottomrule
\end{tabular}
}
\label{tab:main}
\end{table*}

\subsection{What features does \ours learn?}

\begin{figure}[!htbp]%
    \centering
    {\includegraphics[width=1\textwidth]{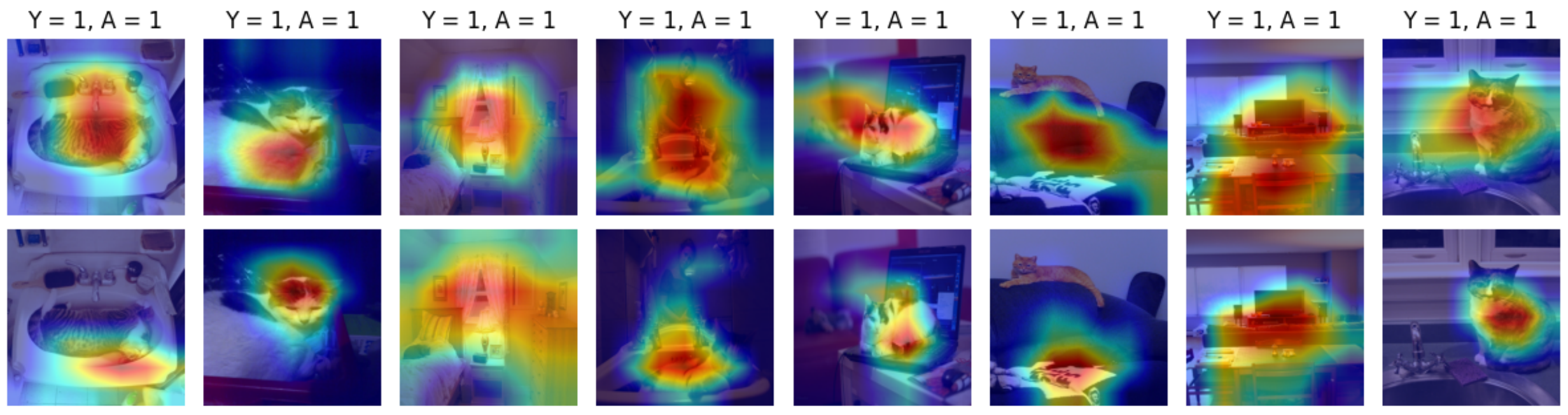} }%
    \caption{We use Grad-CAM to explain the ResNet-18 SSL-base (top), and SSL-\ours model (bottom) for majority examples}%
    \centering
    \label{fig:gcam-maj}
\end{figure}

\begin{figure}[!htbp]%
    \centering
    {\includegraphics[width=1\textwidth]{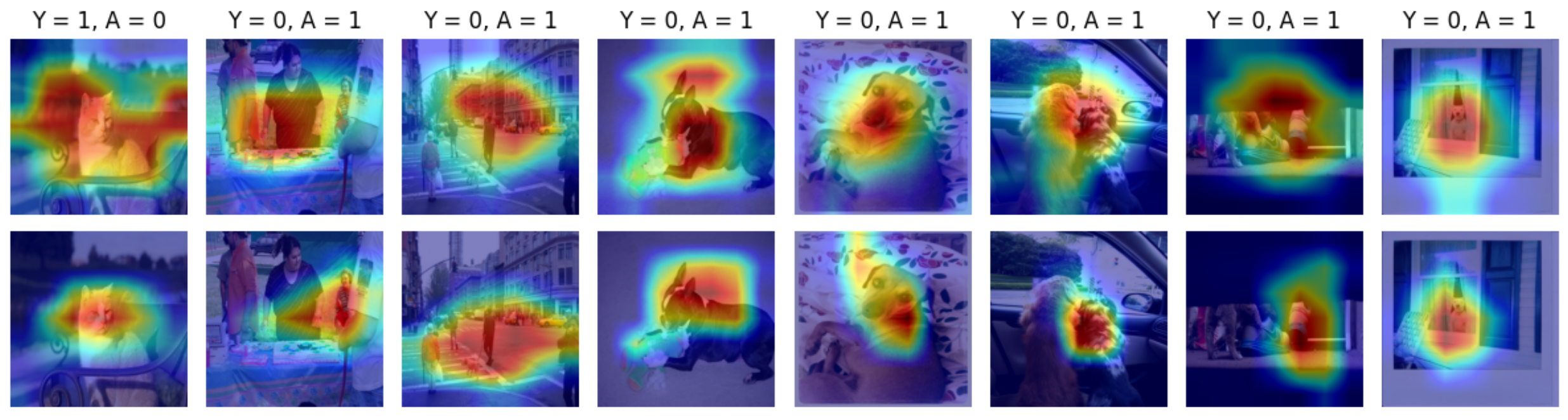} }%
    \caption{We use Grad-CAM to explain the ResNet-18 SSL-base (top), and SSL-\ours model (bottom) for minority examples}
    \centering
    \label{fig:gcam-min}
\end{figure}

Recall that we motivated \ours by explaining that more difficult features could be learned in the \emph{later} layers of an encoder, and by removing the spurious feature from the encoder, we force the model to learn more complex features. In this section, we use Grad-CAM~\cite{selvaraju2016grad} to compare the SSL-base and SSL-\ours. We consider the representations that SSL-base and SSL-\ours learn for \ds{metashift}, and use that to visualize the final layer of the encoder. We choose the best-performing \ours model based on downstream worst-group accuracy. We visualize the parts of the image that both SSL-Base and \ours attend to, in majority~\ref{fig:gcam-maj}, and minority~\ref{fig:gcam-min} groups. 

\subsection{Additional Downstream Imbalance Results}
\label{appdx:minority}

\begin{figure}[h]%
    \centering
    {\includegraphics[width=0.4\textwidth]{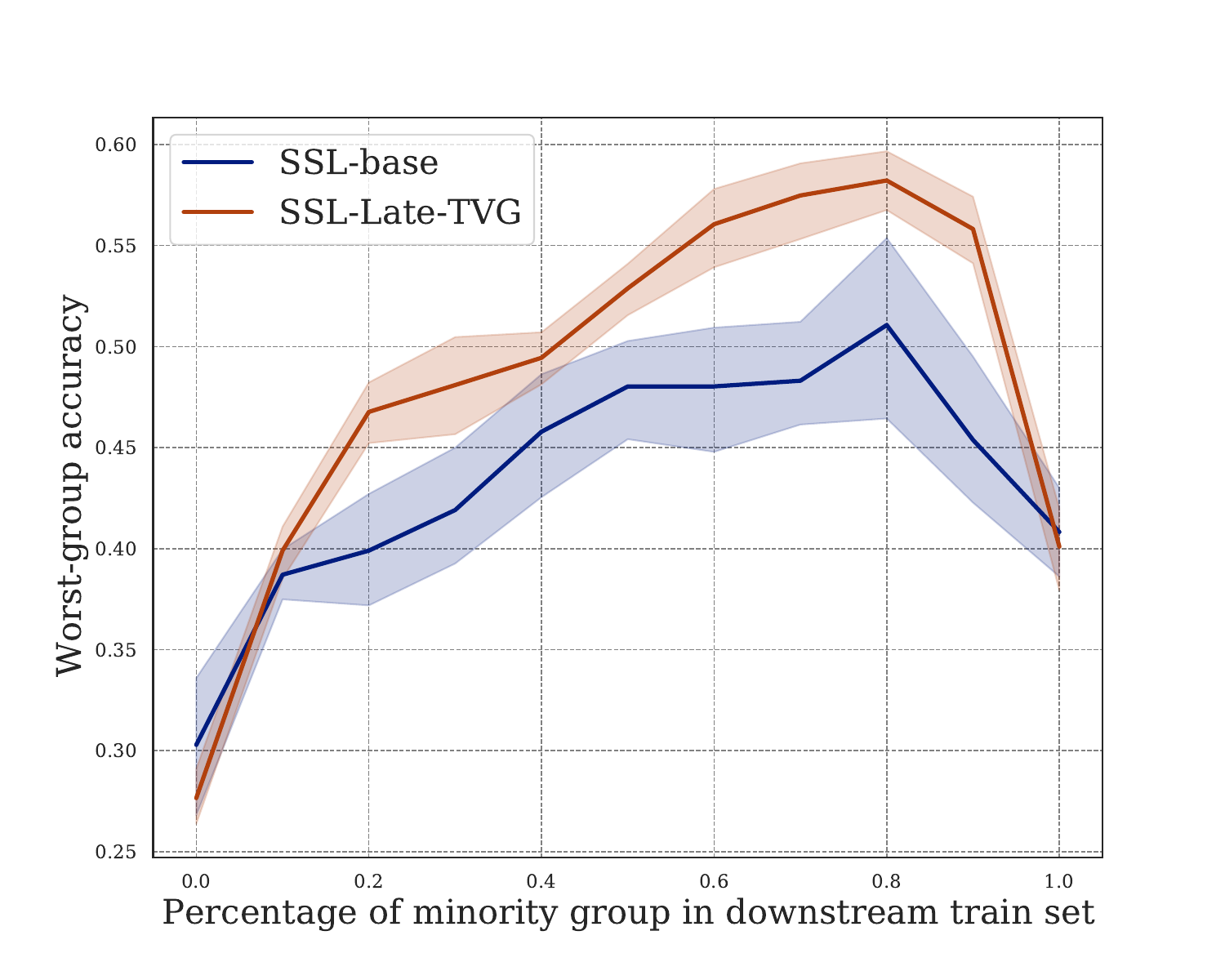} }%
    \qquad
    \centering
    {\includegraphics[width=0.4\textwidth]{figures/min_metashift.pdf} }%
    \caption{Effect of changing minority weight in downstream training set on \ds{metashift}. 
    Left (no regularization), Right (Downstream hyperparameters tuned)}%
    \centering
    \label{fig:imbalance_app}
\end{figure}

For both the best downstream linear model chosen based on worst-group accuracy, and linear models with no regularization, we observe the same trend for the datasets shown in Figure~\ref{fig:imbalance_app}. 

\section{Spurious Learning in Self-supervised Represetations}
\subsection{Additional Re-sampling Results}

We present the complete table from experiment in section~\ref{sec:resample}.

\begin{table}[h]
\centering
\caption{Downstream performance Accuracy (\%) of linear models; For each dataset, we pre-train the model on up-sampled, down-sampled, and balanced training sets}
\begin{tabular}{@{}ccrr@{}}
\toprule
\textbf{Dataset}                      &  \textbf{SSL Train Set} & \textbf{Average} & \textbf{Worst Group} \\ \midrule
\multirow{3}{*}{\ds{celebA}}      & Balanced               & 86.4                         & 75.8                        \\
                                      & Downsampled            & 83.2                         & 77.8                        \\
                                      & Original               & 81.9                         & 77.5                        \\ 
                                      & Upsampled               & 86.3                         & 81.6                        \\ \midrule
\multirow{4}{*}{\ds{cmnist}}      & Balanced               & 75.4                         & 72.0                        \\
                                      & Downsampled            & 74.7                         & 70.1                        \\
                                      & Original               & 82.1                         & 80.7                        \\
                                      & Upsampled              & 77.7                         & 75.4                        \\ \midrule
\multirow{4}{*}{\ds{metashift}}   & Balanced               & 60.7                         & 38.5                        \\
                                      & Downsampled            & 55.7                         & 46.2                        \\
                                      & Original               & 55.8                         & 42.3                        \\
                                      & Upsampled              & 64.4                         & 45.1                        \\ \midrule
\multirow{4}{*}{\ds{spurcifar10}} & Balanced               & 68.7                         & 35.2                        \\
                                      & Downsampled            & 53.1                         & 29.0                        \\
                                      & Original               & 75.0                         & 43.4                        \\
                                      & Upsampled              & 57.4                         & 24.1                        \\ \midrule
\multirow{4}{*}{\ds{waterbirds}}  & Balanced               & 53.1                         & 51.3                        \\
                                      & Downsampled            & 51.0                         & 48.8                        \\
                                      & Original               & 50.7                         & 48.3                        \\
                                      & Upsampled              & 55.2                         & 48.0                        \\ \bottomrule
\end{tabular}
\label{tab:sampling_strat_full}
\end{table}

\subsection{ImageNet Pre-trained Self-supervised Models}\label{app:imagenet}

We obtain pre-trained ResNet50 encoders with SimSiam, SimCLR, and CLIP training strategies, and evaluate the accuracy of core feature prediction similar to the previous sections. 
\begin{table}[!htbp]
\centering
\caption{Worst-group Accuracy (\%) of ImageNet pre-trained models when evaluated on each dataset using a linear probe.}
\begin{tabular}{@{}lrrrr@{}}
\toprule
\textbf{Dataset}       & \multicolumn{1}{r}{\textbf{CLIP}} & \multicolumn{1}{r}{\textbf{ERM}\textsubscript{in}} & \multicolumn{1}{r}{\textbf{SimCLR}\textsubscript{in}} & \multicolumn{1}{r}{\textbf{SimSiam\textsubscript{in}}} \\ \midrule
\ds{celebA}       & 87.2                               & 79.4                                 & 87.2                                    & 84.4                                     \\
\ds{cmnist}       & 88.9                               & 85.4                                 & 85.5                                    & 86.4                                     \\
\ds{metashift}    & 83.1                               & 83.1                                 & 79.7                                    & 69.5                                     \\ 
\ds{waterbirds}   & 81.1                               & 84.3                                 & 78.8                                    & 79.3                    \\
\bottomrule
\end{tabular}
\end{table}

\end{document}